\documentclass[journal]{IEEEtran}
\usepackage{amsmath,amsthm,amssymb}
\usepackage{mathrsfs,mathtools,mathabx}
\usepackage{bm}

\usepackage{dsfont,enumitem,stackrel}
\usepackage{color}
\usepackage{array}
\usepackage{cases}
\usepackage{algorithm,algpseudocode}
\usepackage{algpseudocode}
\usepackage{pgfplots}
\pgfplotsset{compat=newest}
\usepackage{accents}
\usepackage{subfigure}
\usepackage{float}
\usepackage{cite}
\usepackage[most]{tcolorbox}
\usepackage{multicol} 
\usepackage{xr}

\usepackage{xparse}
\usepackage{xcolor}
\usepackage{tikz}
\usetikzlibrary{spy}
\usetikzlibrary{arrows,decorations.pathreplacing,patterns, shapes, positioning,calc}
\usepgfplotslibrary{groupplots}
\usepackage{booktabs} 
\usepackage[nolist,withpage]{acronym}
\usepackage{etoolbox}
\makeatletter
\newif\if@in@acrolist
\AtBeginEnvironment{acronym}{\@in@acrolisttrue}
\newrobustcmd{\LU}[2]{\if@in@acrolist#1\else#2\fi}

\newcommand{\ACF}[1]{{\@in@acrolisttrue\acf{#1}}}

\definecolor{palecgray}{rgb}{0.93, 0.94, 0.94}
\definecolor{chestnut}{rgb}{0.8, 0.36, 0.36}
\definecolor{airforceblue}{rgb}{0.36, 0.54, 0.66}
\definecolor{cadmiumorange}{rgb}{0.93, 0.53, 0.18}
\definecolor{bleudefrance}{rgb}{0.19, 0.55, 0.91}
\definecolor{carolinablue}{rgb}{0.6, 0.73, 0.89}
\definecolor{blue(ncs)}{rgb}{0.0, 0.53, 0.74}
\definecolor{dodgerblue}{rgb}{0.12, 0.56, 1.0}
\definecolor{cssgreen}{rgb}{0.0, 0.5, 0.0}
\definecolor{cadmiumgreen}{rgb}{0.0, 0.42, 0.24}
\definecolor{cadmiumorange}{rgb}{0.93, 0.53, 0.18}
\definecolor{amaranth}{rgb}{0.9, 0.17, 0.31}
\definecolor{bluegray}{rgb}{0.4, 0.6, 0.8}
\definecolor{cadmiumgreen}{rgb}{0.0, 0.42, 0.24}
\definecolor{amber}{rgb}{1.0, 0.75, 0.0}
\definecolor{azure}{rgb}{0.0, 0.5, 1.0}
\definecolor{babyblue}{rgb}{0.54, 0.81, 0.94}
\definecolor{bazaar}{rgb}{0.6, 0.47, 0.48}
\definecolor{celestialblue}{rgb}{0.29, 0.59, 0.82}
\definecolor{darklavender}{rgb}{0.45, 0.31, 0.59}
\definecolor{bluebell}{rgb}{0.64, 0.64, 0.82}
\definecolor{chamoisee}{rgb}{0.63, 0.47, 0.35}
\definecolor{darkcerulean}{rgb}{0.03, 0.27, 0.49}
\definecolor{iris}{rgb}{0.35, 0.31, 0.81}
\definecolor{jazzberryjam}{rgb}{0.65, 0.04, 0.37}
\definecolor{darkcerulean}{rgb}{0.03, 0.27, 0.49}

\newtheorem{prop}{Proposition}
\newtheorem{theorem}{Theorem}
\newtheorem{remark}{Remark}	
\newtheorem{lem}{Lemma}	
\newtheorem{cor}{Corollary}
\theoremstyle{definition}

\def\change{\color{black}}

\definecolor{main}{HTML}{5989cf}    
\definecolor{sub}{HTML}{cde4ff}     

\tcbset{
    sharp corners,
    colback = white,
    before skip = 0.2cm,    
    after skip = 0.5cm      
}

\begin{document}
\bstctlcite{IEEEexample:BSTcontrol}
\begin{acronym}[LTE-Advanced]
  \acro{2G}{second generation}
  \acro{ASK}{amplitude-shift keying}
  \acro{SumComp}{sum computation}
  \acro{6G}{sixth generation}
  \acro{3-DAP}{3-dimensional assignment problem}
  \acro{AA}{antenna array}
  \acro{AC}{admission control}
  \acro{AD}{attack-decay}
  \acro{ADC}{analog-to-digital conversion}
  \acro{ADMM}{alternating direction method of multipliers}
  \acro{ADSL}{asymmetric digital subscriber line}
  \acro{AHW}{alternate hop-and-wait}
  \acro{AI}{artificial intelligence}
  \acro{AirComp}{over-the-air computation}
  \acro{AM}{amplitude modulation}
  \acro{AMC}{adaptive modulation and coding}
  \acro{AP}{\LU{A}{a}ccess \LU{P}{p}oint}
  \acro{APA}{adaptive power allocation}
  \acro{ARMA}{autoregressive moving average}
  \acro{ARQ}{\LU{A}{a}utomatic \LU{R}{r}epeat \LU{R}{r}equest}
  \acro{ATES}{adaptive throughput-based efficiency-satisfaction trade-off}
  \acro{AWGN}{additive white Gaussian noise}
  \acro{BAA}{\LU{B}{b}roadband \LU{A}{a}nalog \LU{A}{a}ggregation}
  \acro{BB}{branch and bound}
  \acro{BCD}{block coordinate descent}
  \acro{BD}{block diagonalization}
  \acro{BER}{bit error rate}
  \acro{BF}{best fit}
  \acro{BFD}{bidirectional full duplex}
  \acro{BLER}{bLock error rate}
  \acro{BPC}{binary power control}
  \acro{BPSK}{binary phase-shift keying}
  \acro{BRA}{balanced random allocation}
  \acro{BS}{base station}
  \acro{BSUM}{block successive upper-bound minimization}
  \acro{CAP}{combinatorial allocation problem}
  \acro{CAPEX}{capital expenditure}
  \acro{CBF}{coordinated beamforming}
  \acro{CBR}{constant bit rate}
  \acro{CBS}{class based scheduling}
  \acro{CC}{congestion control}
  \acro{CDF}{cumulative distribution function}
  \acro{CDMA}{code-division multiple access}
  \acro{CE}{\LU{C}{c}hannel \LU{E}{e}stimation}
  \acro{CL}{closed loop}
  \acro{CLPC}{closed loop power control}
  \acro{CML}{centralized machine learning}
  \acro{CNR}{channel-to-noise ratio}
  \acro{CNN}{\LU{C}{c}onvolutional \LU{N}{n}eural \LU{N}{n}etwork}
  \acro{CP}{computation point}
  \acro{CPA}{cellular protection algorithm}
  \acro{CPICH}{common pilot channel}
  \acro{CoCoA}{\LU{C}{c}ommunication efficient distributed dual \LU{C}{c}oordinate \LU{A}{a}scent}
  \acro{CoMAC}{\LU{C}{c}omputation over \LU{M}{m}ultiple-\LU{A}{a}ccess \LU{C}{c}hannels}
  \acro{CoMP}{coordinated multi-point}
  \acro{CQI}{channel quality indicator}
  \acro{CRM}{constrained rate maximization}
  \acro{CRN}{cognitive radio network}
  \acro{CS}{coordinated scheduling}
  \acro{CSI}{\LU{C}{c}hannel \LU{S}{s}tate \LU{I}{i}nformation}
  \acro{CSMA}{\LU{C}{c}arrier \LU{S}{s}ense \LU{M}{m}ultiple \LU{A}{a}ccess}
  \acro{CUE}{cellular user equipment}
  \acro{D2D}{device-to-device}
  \acro{DAC}{digital-to-analog converter}
  \acro{DC}{direct current}
  \acro{DCA}{dynamic channel allocation}
  \acro{DE}{differential evolution}
  \acro{DFT}{discrete Fourier transform}
  \acro{DIST}{distance}
  \acro{DL}{downlink}
  \acro{DMA}{double moving average}
  \acro{DML}{distributed machine learning}
  \acro{DMRS}{demodulation reference signal}
  \acro{D2DM}{D2D mode}
  \acro{DMS}{D2D mode selection}
  \acro{DNN}{deep neural network}
  \acro{DPC}{dirty paper coding}
  \acro{DRA}{dynamic resource assignment}
  \acro{DSA}{dynamic spectrum access}
  \acro{DSGD}{\LU{D}{d}istributed \LU{S}{s}tochastic \LU{G}{g}radient \LU{D}{d}escent}
  \acro{DSM}{delay-based satisfaction maximization}
  \acro{ECC}{electronic communications committee}
  \acro{EFLC}{error feedback based load control}
  \acro{EI}{efficiency indicator}
  \acro{eNB}{evolved Node B}
  \acro{EPA}{equal power allocation}
  \acro{EPC}{evolved packet core}
  \acro{EPS}{evolved packet system}
  \acro{E-UTRAN}{evolved universal terrestrial radio access network}
  \acro{ES}{edge server}
  \acro{ED}{edge device}
  \acro{FC}{\LU{F}{f}usion \LU{C}{c}enter}
  \acro{FSK}{frequency-shift keying}
  \acro{FD}{\LU{F}{f}ederated \LU{D}{d}istillation}
  \acro{FDD}{frequency division duplex}
  \acro{FDM}{frequency division multiplexing}
  \acro{FDMA}{\LU{F}{f}requency \LU{D}{d}ivision \LU{M}{m}ultiple \LU{A}{a}ccess}
  \acro{FedAvg}{\LU{F}{f}ederated \LU{A}{a}veraging}
  \acro{FER}{frame erasure rate}
  \acro{FF}{fast fading}
  \acro{FL}{federated learning}
  \acro{FEEL}{federated edge learning}
  \acro{FSB}{fixed switched beamforming}
  \acro{FST}{fixed SNR target}
  \acro{FTP}{file transfer protocol}
  \acro{GA}{genetic algorithm}
  \acro{GBR}{guaranteed bit rate}
  \acro{GD}{gradient descent}
  \acro{GLR}{gain to leakage ratio}
  \acro{GOS}{generated orthogonal sequence}
  \acro{GPL}{GNU general public license}
  \acro{GRP}{grouping}
  \acro{HARQ}{hybrid automatic repeat request}
  \acro{HD}{half-duplex}
  \acro{HMS}{harmonic mode selection}
  \acro{HOL}{head of line}
  \acro{HSDPA}{high-speed downlink packet access}
  \acro{HSPA}{high speed packet access}
  \acro{HTTP}{hypertext transfer protocol}
  \acro{ICMP}{internet control message protocol}
  \acro{ICI}{intercell interference}
  \acro{ID}{identification}
  \acro{IETF}{internet engineering task force}
  \acro{ILP}{integer linear program}
  \acro{JRAPAP}{joint RB assignment and power allocation problem}
  \acro{UID}{unique identification}
  \acro{IID}{\LU{I}{i}ndependent and \LU{I}{i}dentically \LU{D}{d}istributed}
  \acro{IIR}{infinite impulse response}
  \acro{ILP}{integer linear problem}
  \acro{IMT}{international mobile telecommunications}
  \acro{INV}{inverted norm-based grouping}
  \acro{IoT}{internet of things}
  \acro{IP}{integer programming}
  \acro{IPv6}{internet protocol version 6}
  \acro{IQ}{in-phase quadrature}
  \acro{ISD}{inter-site distance}
  \acro{ISI}{inter symbol interference}
  \acro{ITU}{international telecommunication union}
  \acro{JAFM}{joint assignment and fairness maximization}
  \acro{JAFMA}{joint assignment and fairness maximization algorithm}
  \acro{JOAS}{joint opportunistic assignment and scheduling}
  \acro{JOS}{joint opportunistic scheduling}
  \acro{JP}{joint processing}
	\acro{JS}{jump-stay}
  \acro{KKT}{Karush-Kuhn-Tucker}
  \acro{L3}{Layer-3}
  \acro{LAC}{link admission control}
  \acro{LA}{link adaptation}
  \acro{LC}{load control}
  \acro{LDC}{\LU{L}{l}earning-\LU{D}{d}riven \LU{C}{c}ommunication}
  \acro{LOS}{line of sight}
  \acro{LP}{linear programming}
  \acro{LTE}{long term evolution}
	\acro{LTE-A}{\ac{LTE}-advanced}
  \acro{LTE-Advanced}{long term evolution advanced}
  \acro{M2M}{machine-to-machine}
  \acro{MAC}{multiple access channel}
  \acro{MANET}{mobile ad hoc network}
  \acro{MC}{modular clock}
  \acro{MCS}{modulation and coding scheme}
  \acro{MDB}{measured delay based}
  \acro{MDI}{minimum D2D interference}
  \acro{MF}{matched filter}
  \acro{MG}{maximum gain}
  \acro{MH}{multi-hop}
  \acro{MIMO}{\LU{M}{m}ultiple-\LU{I}{i}nput \LU{M}{m}ultiple-\LU{O}{o}utput}
  \acro{MINLP}{mixed integer nonlinear programming}
  \acro{MIP}{mixed integer programming}
  \acro{MISO}{multiple input single output}
  \acro{ML}{machine learning}
  \acro{MLWDF}{modified largest weighted delay first}
  \acro{MME}{mobility management entity}
  \acro{MMSE}{minimum mean squared error}
  \acro{MOS}{mean opinion score}
  \acro{MPF}{multicarrier proportional fair}
  \acro{MRA}{maximum rate allocation}
  \acro{MR}{maximum rate}
  \acro{MRC}{maximum ratio combining}
  \acro{MRT}{maximum ratio transmission}
  \acro{MRUS}{maximum rate with user satisfaction}
  \acro{MS}{mode selection}
  \acro{MAE}{\LU{M}{m}ean \LU{A}{a}bsolute \LU{E}{e}rror}
  \acro{MSE}{\LU{M}{m}ean \LU{S}{s}quared \LU{E}{e}rror}
  \acro{MSI}{multi-stream interference}
  \acro{MTC}{machine-type communication}
  \acro{MTSI}{multimedia telephony services over IMS}
  \acro{MTSM}{modified throughput-based satisfaction maximization}
  \acro{MU-MIMO}{multi-user multiple input multiple output}
  \acro{MU}{multi-user}
  \acro{NAS}{non-access stratum}
  \acro{NB}{Node B}
  \acro{NCL}{neighbor cell list}
  \acro{NLP}{nonlinear programming}
  \acro{NLOS}{non-line of sight}
  \acro{NMSE}{normalized mean square error}
  \acro{NN}{neural network}
  \acro{NOMA}{\LU{N}{n}on-\LU{O}{o}rthogonal \LU{M}{m}ultiple \LU{A}{a}ccess}
  \acro{NORM}{normalized projection-based grouping}
  \acro{NP}{non-polynomial time}
  \acro{NRT}{non-real time}
  \acro{NSPS}{national security and public safety services}
  \acro{O2I}{outdoor to indoor}
  \acro{OAC}{over-the-air computation}
  \acro{OFDMA}{\LU{O}{o}rthogonal \LU{F}{f}requency \LU{D}{d}ivision \LU{M}{m}ultiple \LU{A}{a}ccess}
  \acro{OFDM}{orthogonal frequency division multiplexing}
  \acro{OFPC}{open loop with fractional path loss compensation}
	\acro{O2I}{outdoor-to-indoor}
  \acro{OL}{open loop}
  \acro{OLPC}{open-loop power control}
  \acro{OL-PC}{open-loop power control}
  \acro{OPEX}{operational expenditure}
  \acro{ORB}{orthogonal random beamforming}
  \acro{JO-PF}{joint opportunistic proportional fair}
  \acro{OSI}{open systems interconnection}
  \acro{PM}{phase modulation}
  \acro{PAM}{pulse-amplitude modulation}
  \acro{PAIR}{D2D pair gain-based grouping}
  \acro{PAPR}{peak-to-average power ratio}
  \acro{P2P}{peer-to-peer}
  \acro{PC}{power control}
  \acro{PCI}{physical cell ID}
  \acro{PDCCH}{physical downlink control channel}
  \acro{PDD}{penalty dual decomposition}
  \acro{PDF}{probability density function}
  \acro{PER}{packet error rate}
  \acro{PF}{proportional fair}
  \acro{P-GW}{packet data network gateway}
  \acro{PL}{pathloss}
  \acro{PLL}{phase-locked loop}
  \acro{PRB}{physical resource block}
  \acro{PROJ}{projection-based grouping}
  \acro{ProSe}{proximity services}
  \acro{PS}{\LU{P}{p}arameter \LU{S}{s}erver}
  \acro{PSO}{particle swarm optimization}
  \acro{PUCCH}{physical uplink control channel}
  \acro{PZF}{projected zero-forcing}
  \acro{QAM}{quadrature amplitude modulation}
  \acro{QoS}{quality of service}
  \acro{QPSK}{quadri-phase shift keying}
  \acro{RAISES}{reallocation-based assignment for improved spectral efficiency and satisfaction}
  \acro{RAN}{radio access network}
  \acro{RA}{resource allocation}
  \acro{RAT}{radio access technology}
  \acro{RATE}{rate-based}
  \acro{RB}{resource block}
  \acro{RBG}{resource block group}
  \acro{REF}{reference grouping}
  \acro{RF}{radio frequency}
  \acro{RLC}{radio link control}
  \acro{RM}{rate maximization}
  \acro{RNC}{radio network controller}
  \acro{RND}{random grouping}
  \acro{RRA}{radio resource allocation}
  \acro{RRM}{\LU{R}{r}adio \LU{R}{r}esource \LU{M}{m}anagement}
  \acro{RSCP}{received signal code power}
  \acro{RSRP}{reference signal receive power}
  \acro{RSRQ}{reference signal receive quality}
  \acro{RR}{round robin}
  \acro{RRC}{radio resource control}
  \acro{RSSI}{received signal strength indicator}
  \acro{RT}{real time}
  \acro{RU}{resource unit}
  \acro{RUNE}{rudimentary network emulator}
  \acro{RV}{random variable}
  \acro{SAC}{session admission control}
  \acro{SCM}{spatial channel model}
  \acro{SC-FDMA}{single carrier - frequency division multiple access}
  \acro{SD}{soft dropping}
  \acro{S-D}{source-destination}
  \acro{SDPC}{soft dropping power control}
  \acro{SDMA}{space-division multiple access}
  \acro{SDR}{semidefinite relaxation}
  \acro{SDP}{semidefinite programming}
  \acro{SER}{symbol error rate}
  \acro{SES}{simple exponential smoothing}
  \acro{S-GW}{serving gateway}
  \acro{SGD}{\LU{S}{s}tochastic \LU{G}{g}radient \LU{D}{d}escent}  
  \acro{SINR}{signal-to-interference-plus-noise ratio}
  \acro{SI}{self-interference}
  \acro{SIP}{Session Initiation Protocol}
  \acro{SISO}{\LU{S}{s}ingle \LU{I}{i}nput \LU{S}{s}ingle \LU{O}{o}utput}
  \acro{SIMO}{Single Input Multiple Output}
  \acro{SIR}{signal to interference ratio}
  \acro{SLNR}{Signal-to-Leakage-plus-Noise Ratio}
  \acro{SMA}{simple moving average}
  \acro{SNR}{\LU{S}{s}ignal-to-\LU{N}{n}oise \LU{R}{r}atio}
  \acro{SORA}{satisfaction oriented resource allocation}
  \acro{SORA-NRT}{satisfaction-oriented resource allocation for non-real time services}
  \acro{SORA-RT}{satisfaction-oriented resource allocation for real time services}
  \acro{SPF}{single-carrier proportional fair}
  \acro{SRA}{sequential removal algorithm}
  \acro{SRS}{sounding reference signal}
  \acro{SU-MIMO}{single-user multiple input multiple output}
  \acro{SU}{single-user}
  \acro{SVD}{singular value decomposition}
  \acro{SVM}{\LU{S}{s}upport \LU{V}{v}ector \LU{M}{m}achine}
  \acro{TCP}{Transmission Control Protocol}
  \acro{TDD}{time division duplex}
  \acro{TDMA}{\LU{T}{t}ime \LU{D}{d}ivision \LU{M}{m}ultiple \LU{A}{a}ccess}
  \acro{TNFD}{three node full duplex}
  \acro{TETRA}{terrestrial trunked radio}
  \acro{TP}{transmit power}
  \acro{TPC}{transmit power control}
  \acro{TTI}{transmission time interval}
  \acro{TTR}{time-to-rendezvous}
  \acro{TSM}{throughput-based satisfaction maximization}
  \acro{TU}{typical urban}
  \acro{UE}{\LU{U}{u}ser \LU{E}{e}quipment}
  \acro{UEPS}{urgency and efficiency-based packet scheduling}
  \acro{UL}{uplink}
  \acro{UMTS}{universal mobile telecommunications system}
  \acro{URI}{uniform resource identifier}
  \acro{URM}{unconstrained rate maximization}
  \acro{VR}{virtual resource}
  \acro{VoIP}{voice over IP}
  \acro{WAN}{wireless access network}
  \acro{WCDMA}{wideband code division multiple access}
  \acro{WF}{water-filling}
  \acro{WiMAX}{worldwide interoperability for microwave access}
  \acro{WINNER}{wireless world initiative new radio}
  \acro{WLAN}{wireless local area network}
  \acro{WMMSE}{weighted minimum mean square error}
  \acro{WMPF}{weighted multicarrier proportional fair}
  \acro{WPF}{weighted proportional fair}
  \acro{WSN}{wireless sensor network}
  \acro{WWW}{world wide web}
  \acro{XIXO}{(single or multiple) input (single or multiple) output}
  \acro{ZF}{zero-forcing}
  \acro{ZMCSCG}{zero mean circularly symmetric complex Gaussian}
\end{acronym}
%

\title{VecComp: Vector Computing  via MIMO Digital Over-the-Air Computation

 	\thanks{S. Razavikia and C. Fischione are with the School of Electrical Engineering and Computer Science KTH Royal Institute of Technology, Stockholm, Sweden (e-mail: sraz@kth.se, carlofi@kth.se). C. Fischione is also with Digital Futures of KTH. }
 	\thanks{José Mairton B. da Silva Jr. is with the Department of Information Technology, Uppsala University, Sweden (email: mairton.barros@it.uu.se).}
        \thanks{S. Razavikia was supported by the Wallenberg AI, Autonomous Systems and Software Program (WASP), and Ericsson Research Foundation.}
   \thanks{The SSF SAICOM project and the SweWIN research center partially supported this work }

}

\author{
Saeed Razavikia,~\IEEEmembership{Member,~IEEE,}
José Mairton Barros Da Silva Junior,~\IEEEmembership{Member,~IEEE,}
Carlo Fischione,~\IEEEmembership{Fellow,~IEEE}
}

\maketitle

\begin{abstract}
Recently, the ChannelComp framework has proposed digital over-the-air computation by designing digital modulations that enable the computation of arbitrary functions. Unlike traditional analog over-the-air computation, which is restricted to nomographic functions, ChannelComp enables a broader range of computational tasks while maintaining compatibility with digital communication systems. This framework is intended for applications that favor local information processing over the mere acquisition of data. However, ChannelComp is currently designed for scalar function computation, while numerous data-centric applications necessitate vector-based computations, and it is susceptible to channel fading. In this work, we introduce a generalization of the ChannelComp framework, called VecComp, by integrating ChannelComp with multiple-antenna technology.  This generalization not only enables vector function computation but also ensures scalability in the computational complexity, which increases only linearly with the vector dimension. As such, VecComp remains computationally efficient and robust against channel impairments, making it suitable for high-dimensional, data-centric applications. We establish a non-asymptotic upper bound on the mean squared error of VecComp, affirming its computation efficiency under fading channel conditions. Numerical experiments show the effectiveness of VecComp in improving the computation of vector functions and fading compensation over noisy and fading multiple-access channels. 
\end{abstract}

\section{Introduction}
\acresetall

Over-the-air computation (OAC) emerges at the confluence of computation and communication domains. OAC facilitates the computation of mathematical functions by leveraging the superposition characteristic of wireless \ac{MAC}~\cite{nazer2007computation}. This leads to a high-rate data aggregation protocol that can potentially be essential for the next generation of communication networks, catering to applications such as the Internet of Things (IoT) and edge machine learning~\cite{hellstrom2022wireless}. 


OAC, initially focused on analog modulation~\cite{goldenbaum2013harnessing,gastpar2008uncoded}, faces challenges with digital modulation due to nonlinearity and lack of digital coding, impacting reliability and compatibility with digital systems. Recent research addresses these issues by developing basic digital OAC design for a specific function \cite{zhu2020one}. Further advancements that we have conducted have established OAC's compatibility with digital channel coding and generalized digital modulation~\cite{saeed2023ChannelComp,razavikia2023SumCode}; our recent results enable reliable computation and aggregation results, enhancing OAC's robustness and efficiency in modern digital communication systems.

In modern wireless networks, \ac{MIMO} has become the basic technology for improving wireless transceiver performance \cite{adjoudani2003prototype}.  Multiple antennas at both the transmitter and receiving sides increase spectral efficiency, expand range, and strengthen link reliability. If a transmitting node only uses a single antenna, in OAC, we can only compute a single function per available communication resources, such as bandwidth or time slots. However, the next generation of wireless networks with MIMO will allow the simultaneous processing of multi-function purposes, such as matrix-based computation. This motivates us to investigate OAC techniques for MIMO systems in MAC. OAC for MIMO can potentially perform multiple calculations simultaneously and reduce errors using spatial diversity. As a result, the time needed for data combination in computation networks can be reduced. This helps meet the low latency requirements of future networks, especially when high mobility is involved. Complex matrix calculations for machine learning and distributed computing could thus be achieved.

\subsection{Literature Review}

In the seminal works by \cite{nazer2007computation,gastpar2008uncoded}, the computing principles across MAC have been systematically studied, focusing on the theoretical limits for a predefined many-to-one function. In \cite{goldenbaum2013harnessing,goldenbaum2014nomographic}, the authors have established a connection between OAC and nomographic functions and expanded the range of functions that OAC can compute, including a set broader than merely linear functions. This connection has shown that OAC offers a superior computation rate compared to isolated communication and computation processes. Due to these promising theoretical findings, the OAC framework has garnered increased attention. The domain of OAC has been investigated from diverse perspectives, such as information theory~\cite{nazer2011compute}, signal processing \cite{goldenbaum2013harnessing}, and synchronization challenges~\cite{goldenbaum2013robust,saeed2022BlindFed,hellstrom2023optimal}. Notably, the analog method OAC has been recognized for its potential to optimize communication resources, especially in federated edge learning~\cite{amiri2020federated}.

However, dependency on analog communication makes the reliability of OAC questionable due to the wireless channel distortions and noise~\cite{csahin2023over}. Moreover, its dependence on analog hardware is limiting, given the few modern devices supporting analog modulations \cite{zhu2019broadband,saeed2023ChannelComp}. Towards devising a digital aggregation method,  one-bit broadband digital aggregation, using simple binary phase shift keying modulation, has been introduced in ~\cite{zhu2020one} with its extension to non-coherent system model studied in~\cite{ csahin2023over,sahin2024over}.   
 However, the highlighted efforts predominantly focus on particular functions and are confined to specific machine learning training methods, such as signSGD~\cite{bernstein2018signsgd}.  Towards boosting reliability, the works in \cite{zhu2018mimo,Zhu2021MassiveIoT} introduce zero-forcing beamforming, which mitigates interference in analog OAC systems. In contrast, \cite{zhai2021hybrid} proposes hybrid beamforming for massive MIMO OAC, leveraging spatial diversity to enhance computational accuracy. A significant advancement is presented in \cite{Jing2023MIMO}, which formulates an optimal transceiver beamforming design for multi-antenna transmitters and receivers. Similarly, \cite{chen2024over} studies OAC in cell-free MIMO systems, analyzing the impact of network cooperation on computational accuracy. These studies demonstrate the potential of MIMO to improve OAC performance but remain constrained to analog transmission, which is inherently susceptible to noise accumulation. Furthermore, their beamforming designs presume perfect CSI at both transmitters and receiver ends and impose stringent synchronization requirements.    

  Most existing OAC frameworks are designed to compute scalar functions, typically nomographic functions that can be expressed as summations of local components. While effective for applications such as federated learning \cite{amiri2020federated} and distributed sensing \cite{Zhu2021MassiveIoT}, these methods are not directly applicable to more general vector function computations, such as nonlinear activation functions in neural network architecture.

In recent works \cite{saeed2023ChannelComp,razavikia2023computing, razavikia2023SumCode,razavikia2024FedComp,razavikia2025designing,razavikia2025designing2}, we have presented the ChannelComp framework, a novel communication for computation paradigm. The ChannelComp framework enables the computation of various functions over the \ac{MAC} utilizing digital modulations. The ChannelComp provides a spectral efficient communication approach via digital modulation for the OAC challenge. Moreover, ChannelComp offers computational advantages over analog OAC techniques and integrates seamlessly with digital communication infrastructures~\cite{perez2024waveforms}. Building on the ChannelComp concept, recent works~\cite{liu2024digital,yan2024novel}, have explored channel coding techniques within digital OAC to bolster communication reliability.

While ChannelComp has pioneered a new avenue for performing computations via communication, it assumes networks with single antennas, thus performing scalar output functions. Furthermore, ChannelComp has not factored in fading channels within its computational framework, potentially leading to unreliable communication and subsequent computation inaccuracies during outage instances.

Here, we aim to generalize ChannelComp's computational capabilities by integrating vector computation by incorporating MIMO design. Including MIMO widens the dimensions of computation in ChannelComp, increases spectral efficiency, and reinforces computation accuracy. 

However, generalizing ChannelComp to support vector function computation in digital OAC over MIMO systems introduces several fundamental challenges. We outline the key challenges as follows:

\begin{itemize}

    \item \textbf{Vector computation in multi-antenna systems:} While conventional OAC handles only scalar aggregation, extending it to vector functions requires new designs that guarantee separability across multiple output dimensions.
    
    \item \textbf{CSI-unaware design:} Many existing digital OAC frameworks rely on full CSI at the transmitters to optimize signal transmission. However, acquiring CSI at distributed nodes introduces significant overhead and delays, making real-time computation challenging. 

    \item \textbf{Scalability in high-dimensional computation:}  As the number of inputs and nodes grows, ChannelComp constraints can increase exponentially, posing challenges to maintaining computational accuracy. 
    
    \item \textbf{Compensation for correlated channels:} In practical multi-antenna systems, channel coefficients often exhibit spatial correlation, deviating from the idealized independent and identically distributed (i.i.d.) assumption. This correlation affects the superposition property of OAC, making it more difficult to achieve reliable function computation.

\end{itemize}

The following section presents our contributions, which address these challenges and provide a structured approach to advancing digital over-the-air computation in MIMO systems.

\subsection{Our Contributions}

This paper proposes extending the ChannelComp scheme, termed VecComp, enabling computing matrix-based functions over a finite field through fading \ac{MAC}s. Notably, the VecComp uses MIMO to provide reliable communication and spectral efficiency with low latency, resulting in a high rate for both matrix- and vector-based function computation. The VecComp framework notably eliminates the need for network nodes to be aware of the \ac{CSI}. This represents a clear evolution from traditional methods used in OAC via a wireless fading \ac{MAC}. Typically, each transmitting node adjusts its transmission based on the real-time channel state to ensure consistent power level convergence at the receiver. By incorporating multiple antennas at the receiver, VecComp diminishes the fading effects intrinsic to communication channels, reinforcing its functional capacity. Additionally, we analyze the theoretical effectiveness of the VecComp framework. An in-depth analysis is studied to ascertain the required antennas at the receiver for effectively nullifying the wireless channel's fading characteristics.

Specifically, our contributions are as follows:

\begin{itemize}
    \item \textbf{MIMO-based digital OAC for vector computation:}  
    We introduce VecComp, which extends scalar OAC to vector functions by decoupling channel compensation from compaction. Its complexity grows only linearly with the number of functions and exploits MIMO to compute multiple outputs simultaneously.
    
    \item \textbf{CSI-unaware OAC computation:}  
    VecComp uses receiver-side CSI and multi-antenna diversity to counter small scale fading, eliminating the need for transmitter CSI and thus reducing delay and boosting spectral efficiency. 
    
    \item \textbf{Handling correlated fading channels:}  
    We quantify how spatial correlation degrades computation accuracy and show that our beamforming strategy effectively mitigates these effects in realistic fading environments.
    
    \item \textbf{Theoretical guarantees:}  We provide the analysis for the required number of antennas $N_r$ at the receiver to obtain an upper bound for the function’s computation error to a certain level $\epsilon$. In particular, we derive a probabilistic lower bound on the required receiver antennas, proving \(N_r=\mathcal{O}(1/\epsilon^2)\) to ensure an error tolerance \(\epsilon\). 

    \item \textbf{Numerical experiments:}  We validate the theoretical results and assess the VecComp framework’s performance
effectiveness via extensive numerical experiments. These experiments provide numerical results regarding the computation performance and the proposed fading compensation. Simulation results confirm the theoretical bounds and show that increasing \(N_r\) can cut computation error by up to $75$\%. 
    
\end{itemize}

Finally, VecComp is a fully digital OAC framework, meaning that it can be directly integrated into modern wireless systems that support digital modulations (e.g., QAM, PSK). This makes it suitable for deployment in {cellular networks}, {IoT applications}, and {edge computing}.

\subsection{Document Organization}

The rest of the paper is organized as follows: in Section~\ref{sec:system}, we explain the system model, including the signal models. Next, we give the problem formulation and the methodology for performing the computation over the \ac{MAC} in Section~\ref{sec:Modulation}. In Section~\ref{sec:FunctionComp}, we introduce the proposed VecComp method for computing multiple functions over the \ac{MAC}. Then, we evaluate the performance of VecComp in terms of MSE over fading and noisy channels in Section~\ref{sec:num}. Finally, we conclude the paper in Section~\ref{sec:conclusion}.

\subsection{Notation}
We denote a finite field by $\mathbb{F}_Q$, where $Q$ denotes the number of elements inside the field. Moreover, we denote by $\mathbb{Z}$,  $\mathbb{R}$, and $\mathbb{C}$ as the integer, real, and complex number sets, respectively. We use lowercase letters $x$ for scalar and calligraphic notation $\mathcal{X}$ to represent operators.  The transpose and Hermitian of a matrix $\bm{X}$ are represented by $\bm{X}^{\mathsf{T}}$ and $\bm{X}^{\mathsf{H}}$, respectively. For a vector $\bm{x}$, $\|\bm{x}\|_1, \|\bm{x}\|_2$  are defined as the  $\ell_1$ and $\ell_2$ vector norms, respectively. We define $\|\bm{X}\|$ and $\|\bm{X}\|_{\rm F}$  as the spectral and Frobenius norms of the matrix $\bm{X}$, respectively. For an integer $N$, $[N]$ corresponds to the set  $\{1,2,\dots, N\}$. We define $\mathcal{W}_f$ as the range of function $f$, and  its cardinality  by $|\mathcal{W}_f|$.  We use $\bm{X}\succeq \bm{0}$ to show that $\bm{X}$ is a positive semidefinite matrix.  Finally, we define the $ {\rm diag}: \mathbb{C}^N \mapsto \mathbb{C}^{N\times N}$ operator over a vector $\bm{x} \in \mathbb{C}^{N}$ as the mapping to a diagonal matrix whose diagonal elements are the entries of vector $\bm{x}$. The null-space of linear operator $\mathcal{X}$ is denoted by $\mathcal{N}(\mathcal{X})$.  We define the linear block diagonal operator \(\mathcal{T}\) as the direct sum of operators $\{\mathcal{X}_i\}_{i=1}^{L}$, denoted by \(\mathcal{T} := \bigoplus_{i=1}^{L}\mathcal{X}_i\), where for a complex vector \(d = (d_1, \ldots, d_L) \in \mathbb{C}^{L}\), the operator \(\mathcal{T}\) is applied component-wise, yielding the output \(\mathcal{T}(d) = [\mathcal{X}_1(d_1), \ldots, \mathcal{X}_L(d_L)]\).

\section{System Model and Problem Formulation}\label{sec:system}

\subsection{System Model}
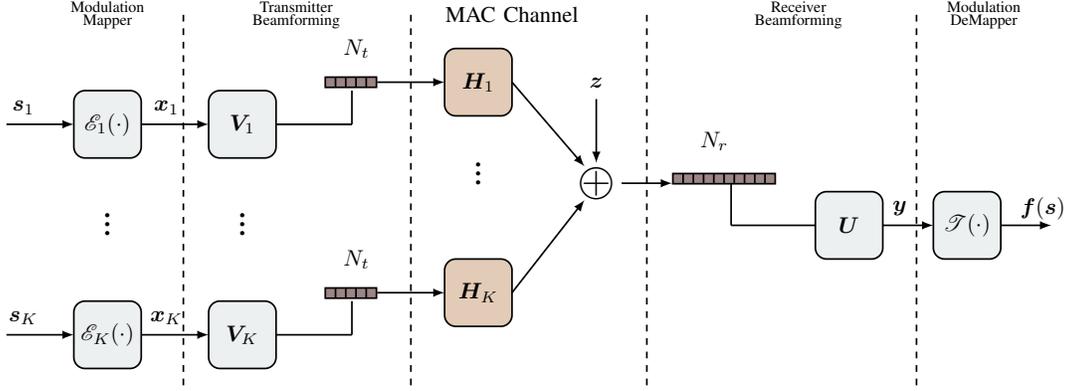
\begin{figure*}
    \centering
    \scalebox{0.85}{
    \tikzset{every picture/.style={line width=0.75pt}}     

\begin{tikzpicture}[x=0.75pt,y=0.75pt,yscale=-1,xscale=1]

\draw (90,0) node     { $\substack{\text{Modulation} \\  \text{Mapper}}$};
\draw (202,0) node    { $\substack{\text{Transmitter} \\  \text{Beamforming}}$};
\draw (330,0) node    { MAC Channel};
\draw (500,0) node   { $\substack{\text{Receiver} \\  \text{Beamforming}}$}; 
\draw (610,0) node    {$\substack{\text{Modulation} \\  \text{DeMapper}}$};

\draw[dashed]    (135,0) -- (135,220) ;
\draw[dashed]    (270,0) -- (270,220) ;
\draw[dashed]    (410,0) -- (410,220) ;
\draw[dashed]    (570,0) -- (570,220) ;

\draw (90, 120) node   {\LARGE $\vdots $};
\draw (170, 120) node   {\LARGE $\vdots $};
\draw (310, 90) node   {\LARGE $\vdots $};

\draw    (190,65) -- (235,65) ;

\draw (40,55) node  {$\bm{s}_{1}$};
\draw (40,180) node  {$\bm{s}_{K}$};

\draw[-latex]    (30,65) -- (70,65) ;
\draw[-latex]    (30,190) -- (70,190) ;

\draw (125,55) node     {$\bm{x}_{1}$};
\draw (125,180) node     {$\bm{x}_{K}$};

\draw[fill=palecgray, rounded corners=5pt] (110, 85) rectangle (70, 45) {};

\draw [fill=palecgray, rounded corners=5pt] (110, 210) rectangle (70, 170) {};

\draw (90,65) node    {$\mathscr{E}_{1}(\cdot)$};
\draw (90,190) node    {$\mathscr{E}_{K}(\cdot)$};

\draw[-latex]    (110,65) -- (150,65) ;
\draw[-latex]    (110,190) -- (150,190) ;

\draw[fill=palecgray, rounded corners=5pt] (190, 85) rectangle (150, 45) {};
\draw (170,65) node    {$\bm{V}_{1}$};

\draw[fill=palecgray, rounded corners=5pt] (190, 210) rectangle (150, 170) {};
\draw (170,190) node    {$\bm{V}_{K}$};

\draw    (190,65) -- (235,65) ;
\draw    (190,190) -- (235,190) ;


\draw (450,75) node  {$N_{r}$};
\draw  [draw opacity=0][fill={rgb, 255:red, 156; green, 133; blue, 133 }  ,fill opacity=1 ] (425.43,94.24) -- (486.18,94.24) -- (486.18,100.31) -- (425.43,100.31) -- cycle ; \draw   (431.51,94.24) -- (431.51,100.31)(437.58,94.24) -- (437.58,100.31)(443.66,94.24) -- (443.66,100.31)(449.73,94.24) -- (449.73,100.31)(455.8,94.24) -- (455.8,100.31)(461.88,94.24) -- (461.88,100.31)(467.95,94.24) -- (467.95,100.31)(474.03,94.24) -- (474.03,100.31)(480.1,94.24) -- (480.1,100.31) ; \draw    ; \draw   (425.43,94.24) -- (486.18,94.24) -- (486.18,100.31) -- (425.43,100.31) -- cycle ;

\begin{scope}[shift={(0.73cm,0cm)}]
\draw (210,20) node   {$N_{t}$};
    \draw  [draw opacity=0][fill={rgb, 255:red, 156; green, 133; blue, 133 }  ,fill opacity=1 ] (191.5,37.01) -- (221.88,36.96) -- (221.89,43.04) -- (191.51,43.08) -- cycle ; \draw   (197.58,37) -- (197.59,43.08)(203.65,36.99) -- (203.66,43.07)(209.73,36.98) -- (209.74,43.06)(215.8,36.97) -- (215.81,43.05) ; \draw    ; \draw   (191.5,37.01) -- (221.88,36.96) -- (221.89,43.04) -- (191.51,43.08) -- cycle ;
\end{scope}

\draw    (235,65) -- (235,45) ;

\begin{scope}[shift={(0.73cm,3.3cm)}]
\draw (210,20) node   {$N_{t}$};
    \draw  [draw opacity=0][fill={rgb, 255:red, 156; green, 133; blue, 133 }  ,fill opacity=1 ] (191.5,37.01) -- (221.88,36.96) -- (221.89,43.04) -- (191.51,43.08) -- cycle ; \draw   (197.58,37) -- (197.59,43.08)(203.65,36.99) -- (203.66,43.07)(209.73,36.98) -- (209.74,43.06)(215.8,36.97) -- (215.81,43.05) ; \draw    ; \draw   (191.5,37.01) -- (221.88,36.96) -- (221.89,43.04) -- (191.51,43.08) -- cycle ;
\end{scope}

\draw    (235,190) -- (235,170) ;

\draw[-latex]    (250,40) -- (290,40) ;
\draw[-latex]    (250,165) -- (290,165) ;

\draw[-latex]    (330,40) -- (373,90) ;

\draw[fill={rgb, 255:red, 194; green, 142; blue, 96 }  , rounded corners=5pt, fill opacity=0.45 ] (330, 60) rectangle (290, 20) {};

\draw (310,40) node  {$\bm{H}_{1}$};

\draw[-latex]    (395,100) -- (425,100) ;

\draw (380,40) node    {$\bm{z}$};
\draw[-latex]    (380,50) -- (380,88) ;

\draw (380,100) node {\Large $\bigoplus$} ;

\draw[fill={rgb, 255:red, 194; green, 142; blue, 96 }  , rounded corners=5pt, fill opacity=0.45 ] (330, 185) rectangle (290, 145) {};

\draw (310,165) node  {$\bm{H}_{K}$};

\draw[-latex]    (330,165) -- (373,110) ;




\draw    (460,100) -- (460,125) ;

\draw    (460,125) -- (510,125) ;

\draw[fill=palecgray , rounded corners=5pt] (550, 145) rectangle (510, 105) {};

\draw[-latex]    (550,125) -- (580,125) ;

\draw (530,125) node    {$\bm{U}$};

\draw (560,115) node   {$\bm{y}$};

\draw[fill=palecgray , rounded corners=5pt] (620, 145) rectangle (580, 105) {};

\draw[-latex]    (620,125) -- (650,125) ;

\draw (645,115) node   {$\bm{f}(\bm{s})$};


\draw (600,125) node    {$\mathscr{T}(\cdot)$};

\end{tikzpicture}
}

    \caption{MIMO Setup with $k$ transmitters, where there are $N_t$ transmitter antenna at nodes and $N_r$ receivers antenna at the CP. Node $k$ uses a beamforming matrix $\bm{V}_k \in \mathbb{C}^{N_t \times L}$ to transmit the signal $\bm{x}_k$ derived from the encoded source signal $\bm{s}_k$ via the encoder $\mathscr{E}_k(\cdot)$.  The signal passes through the channel matrix $\bm{H}_k$ and is combined into the additive noise, $\bm{z}$, thus representing the aggregated received signal. The received signal is then processed by the beamforming matrix $\bm{U}$ to produce the final output $\bm{y}$.}
    \label{fig:MIMO-setup}
\end{figure*}

We have a communication system involving a \ac{CP} with $N_r$ receiver antennas. Also, there are $K$ nodes with $N_t$ transmitter antennas and a single receiver antenna. The nodes communicate with the \ac{CP} using a shared communication channel. We consider that node $k$ owns value $s_{k}\in \mathbb{F}_{Q}$, where $Q$ denotes the number of elements inside the input domain field $\mathbb{F}_{Q}$. The objective of the \ac{CP} is to compute the desired function $\bm{f}({s}_1,\ldots,{s}_K):\mathbb{F}_{Q}^{K}\mapsto \mathbb{R}^{L}$ or, $\bm{f}(\bm{s})$  in short, with $L$ outputs.  Note that $L$ must be lower than the number of receiver and transmitter antennas, i.e., $L\leq \min\{N_r, N_t\}$\footnote{ This condition is unavoidable for perfect recovery of $\sum_{k}\bm{x}_k$ over MAC because it is a fundamental limitation dictated by MIMO linear algebra, regardless of the $\bm{x}_k$ values. Nevertheless, by adopting wideband transmission over multiple time slots or frequency bands~\cite{yan2025remac,yan2024novel}, it is possible to increase the number of computational streams beyond the constraints imposed by the per-device antenna count. When bandwidth is restricted, a wait-and-transmit strategy~\cite{shao2013distributed} can be adopted to alleviate this constraint.
}. In the notable case of $L=1$, the computation problem would be reduced to the ChannelComp~\cite{saeed2023ChannelComp} scalar function problem. Therefore, in this paper, we extend the computation problem to a vector function instead of a scalar function, using beamformers at transmitters and receivers to decrease distortion. 
As an illustrative example,  we present the following affine function $\bm{f}$ for the given input vector $\bm{s} \in \mathbb{F}_{Q}^{K}$, 
\begin{align}
        \bm{f}(\bm{s}) = \bm{A}\bm{s} + \bm{b} \in \mathbb{R}^{L},  
\end{align}
where $\bm{A}\in \mathbb{R}^{L\times K}, \bm{b} \in \mathbb{R}^{L}$. Here, element $k$ of the input vector, ${s}_k$, corresponds to data of node $k$, for $k\in [K]$; and element $\ell$ of the output vector, $[\bm{f}]_{\ell}$, corresponds to the data received by antenna $\ell$ at the receiver, accordingly.
 
 Throughout digital transmission process,  the value $s_{k}$ at node $k$ is mapped to $L$ different digitally modulated signal $x_{k,\ell}$ via the encoder $\mathscr{E}_{k,\ell}(\cdot)$ such that ${x}_{k,\ell}= \mathscr{E}_{k,\ell}({s}_{k})$, for $\ell \in [L]$. Let $\bm{V}_k\in \mathbb{C}^{N_t\times L}$ be the transmit beamforming matrix at node $k$ and let $\bm{U}\in \mathbb{C}^{N_r\times L}$ be the receiver beamforming matrix (see Figure~\ref{fig:MIMO-setup}). Then, the nodes transmit their respective $\bm{x}_k$ values, where $\bm{x}_k = [x_{k,1},\ldots,x_{k,\ell}]^{\mathsf{T}}\in \mathbb{C}^{L}$, using beamforming matrix $\bm{V}_k$  over \ac{MAC} to compute the function $\bm{f}$ at the \ac{CP}. In particular, assuming symbol-level synchronization, all users transmit their symbol vectors simultaneously using their arrays.  Then, the measured signal received by the \ac{CP} over the \ac{MAC} is given by
\begin{align}
    \label{eq:MIMO-setup-1}
      \bm{y} = \bm{U}^{\mathsf{H}}\Big(\sum\nolimits_{k=1}^K \frac{\bm{H}_k}{N_t}\bm{V}_k\bm{x}_k+ \bm{z}\Big),
\end{align}
where  $\bm{z} \in \mathbb{C}^{N_r}$ denotes the AWGN vector with i.i.d.  distribution from $\mathcal{CN}(0,\sigma_z^2 \bm{I}_{N_r})$, and $\bm{H}_k \in \mathbb{C}^{N_r \times N_t}$  denotes the MIMO channel matrix for the link from node $k$ to the \ac{CP}, whose entries are generated i.i.d from complex Normal distribution, i.e. $\bm{H}_k\sim \mathcal{CN}(\bm{0},\sigma_{h,k}^2\bm{I}_{N_r})$.  The distortion of the received vector concerning the target function vector due to the MIMO channel and noise is suppressed using transmit and receive beamforming. In other words, joint beam formation attempts to combine $K$ symbol vectors in the \ac{CP} coherently. To compute the function $\bm{f}$,  we use an operator $\mathscr{T}$ to map the resultant constellation diagram of signal $\bm{y}$ to the range of the function $\bm{f}$, i.e., $\mathscr{T}(\bm{y}) = \bm{f}(\bm{s})$, where $\bm{s}:=[s_1,\ldots, s_K]^{\mathsf{T}}\in \mathbb{F}_{Q}^{K}$.

Note that each component within the $\bm{y}$ constellation diagram is influenced by three processes: the summation given in~\eqref{eq:MIMO-setup-1}, the number of the nodes $K$, and the selected modulations for each $\bm{x}_k$. Note that other factors impact the constellation diagram of $\bm{y}$, such as the transmitter beamforming $\bm{V}_k$ and the receiver beamforming matrix $\bm{U}$. Consequently, the resulting constellation diagram represents a transformation of the transmitting nodes' original constellation diagram, denoted as $\bm{x}_k$, modulated by the receiver beamforming matrix $\bm{U}$ operations.

\begin{remark}\label{rem_nonideail}
Several non-idealities must be considered in practical implementations of VecComp, including synchronization errors, carrier frequency offset,  multipath fading, channel estimation inaccuracies at the CP, and noise distribution~\cite{perez2024waveforms}. These factors can impact the accuracy of OAC by introducing phase and amplitude misalignments in the aggregated signal. However, various techniques, such as non-coherent aggregation~\cite{saeed2022BlindFed}, retransmission strategy \cite{yan2025remac}, and bit slicing \cite{liu2024digital}, can mitigate these effects.  VecComp exhibits inherent robustness against such errors due to its beamforming-based design and the randomization introduced in the transmission process. The randomized beamforming technique, which will be discussed in Section~\ref{sec:CSI-Unaware_beam}, 
ensures that small synchronization mismatches do not destructively affect the computation accuracy, making VecComp a practical and resilient solution for real-world wireless computation systems.
\end{remark}

\subsection{Problem Formulation}

If we can appropriately choose the modulation vectors $\bm{x}_1,\ldots,\bm{x}_K$,  the mapping $\mathscr{T}(\bm{y})$ can be determined based on the diagram of the resultant constellation points by $\sum_{k}\bm{x}_k$ such that its output approximates the value of function $\bm{f}$. Hence, to obtain the sought modulation vectors $\bm{x}_1,\ldots,\bm{x}_K$, we propose the following optimization problem.
\begin{align}
\nonumber
	\mathcal{P}_{0} :=  \underset{\substack{\bm{U},\bm{V}_1,\ldots,\bm{V}_K\\\bm{x}_1,\ldots,\bm{x}_K}}{\rm min}  \sum\nolimits_{\bm{s} \in \mathcal{D}_f }& \Big\|\bm{f}(\bm{s})-\mathscr{T}\big(\bm{y}\big)\Big\|_2^2, \\ 	\label{eq:Associate}
 \text{s.t.}~ &  \frac{\|\bm{V}_k\bm{x}_k\|_2}{N_t} \leq P_{\rm max},~ k\in [K], 
\end{align}
where $P_{\rm max}$ is the available power budget at the nodes, and $\mathcal{D}_f$ is the domain of function $\bm{f}$.   The bilinear product  $\bm{V}_k\bm{x}_k$ and nonlinear operators $\mathscr{E}_k$ make Problem $\mathcal{P}_0$ non-convex and highly challenging. To solve this problem, we use a \textit{separation scheme} to design beamforming matrices at the transmitter and receiver, thus helping to compensate for the communication channel's fading and noise effects. Subsequently, we propose the encoders $\mathscr{E}_k$ accordingly to the decoder  $\mathscr{T}$ to perform the vector computation through \ac{OAC} in digital MIMO communications. In the next section, we proceed with the methodology for the fading channel compensation.

\section{Statistical Channel Compensation }\label{sec:Modulation}

 In this section, we design the beamforming matrices for a scenario where the transmitter nodes are unaware of \ac{CSI}. Consequently, we use the zero-forcing technique at the \ac{CP} and consider the asymptotic massive MIMO phenomena for compensating the channel effects. Then, we analyze the efficiency of such a fading channel compensation in terms of the mean square of the computation error.  
 
\subsection{CSI-Unaware Fading Compensation via Beamforming}\label{sec:CSI-Unaware_beam}

Without loss of generality, we assume the modulated signals are normalized, i.e., $\|\bm{x}_k\|_2^2 =1,$ for $k\in[K]$.  Consequently, from the power constraints in~\eqref{eq:Associate}, the norm of the beamforming matrix $\bm{V}_k$ is constrained by the power budget $P_{\max}$, such that $\|\bm{V}_k\| \leq P_{\rm max}N_t$ for all $k \in [K]$. To compensate for the fading effect, all the nodes generate the beamforming matrices  $\bm{V}_{k}$'s according to a given distribution $\mathcal{F}_k$ obeying the isotropy and incoherence properties discussed shortly. Assume that the \ac{CP} has access to the perfect \ac{CSI} and the beamforming matrices for all the $K$ nodes, i.e., $\bm{H}_k$ and $\bm{V}_k$ for $k \in [K]$. Regardless of distributions $\mathcal{F}_k$s, the  \ac{CP} can set $\bm{U} = \sum_{k=1}^K\bm{V}_k^{\mathsf{H}}\bm{H}_k^{\mathsf{H}}/N_r$, which yields the following equation:
\begin{align}
    \nonumber
      \bm{y} = & \frac{1}{\beta}\sum_{k=1}^K\bm{V}_k^{\mathsf{H}}\bm{H}_k^{\mathsf{H}}\bm{H}_k\bm{V}_k\bm{x}_k +  \frac{1}{\beta}\sum_{k,k',k\neq k' }^K\bm{V}_{k'}^{\mathsf{H}}\bm{H}_{k'}^{\mathsf{H}}\bm{H}_k\bm{V}_k\bm{x}_k  \\ \label{eq:MIMO-setup-2}
      & + \frac{1}{\beta} \sum\nolimits_{k=1}^K\bm{V}_k^{\mathsf{H}}\bm{H}_k^{\mathsf{H}}\bm{z},
\end{align}
where we define $\beta = N_rN_t$ as the normalizing factor. Then, we can rewrite~\eqref{eq:MIMO-setup-2} as follows: $\bm{y} = \bm{y}_{\rm sig} + \bm{y}_{\rm inter} + \bm{y}_{\rm noise},$ where the signal terms belong to $\mathbb{C}^L$ and are given by 
\begin{subequations}
\begin{align}
     \bm{y}_{\rm sig} & := \frac{1}{\beta}\sum\nolimits_{k=1}^K\bm{V}_k^{\mathsf{H}}\bm{H}_k^{\mathsf{H}}\bm{H}_k\bm{V}_k\bm{x}_k,\\
      \bm{y}_{\rm inter} & :=  \frac{1}{\beta}\sum\nolimits_{k,k',k\neq k' }^K\bm{V}_{k'}^{\mathsf{H}}\bm{H}_{k'}^{\mathsf{H}}\bm{H}_k\bm{V}_k\bm{x}_k,  \label{eq:signalinter}\\
       \bm{y}_{\rm noise}& := \frac{1}{\beta} \sum\nolimits_{k=1}^K\bm{V}_k^{\mathsf{H}}\bm{H}_k^{\mathsf{H}}\bm{z}. 
\end{align}
\end{subequations}
Here,  $\bm{V}_k$ is  generated according to some distribution $\mathcal{F}_k$ for $k\in[K]$.  To reduce the variance of the interference term $\bm{y}_{\rm inter}$ and increase the power of signal term $\bm{y}_{\rm sig}$, we consider the distribution $\mathcal{F}_k$ to obey the \textit{isotropy} and \textit{independency} properties~\cite{daei2024timely,heckel2017generalized}. Specifically, we assume that the distribution of $\bm{V}_k$s generated from $\mathcal{F}_k$ satisfy the following properties: 
\begin{align}
    \label{eq:Isotropy}
    & \textit{Isotropy property:}~\mathbb{E}[\bm{V}_k^{\mathsf{H}}\bm{V}_k] = \sigma_{v,k}^2N_t\bm{I}_{L},~\bm{V}_k \sim \mathcal{F}_k,\\  \label{eq:Independency}
    & \textit{Independency property:}~\mathbb{E}[\bm{V}_k^{\mathsf{H}}\bm{V}_{k'}]  = \bm{0}_{L},~ \bm{V}_{k'} \sim \mathcal{F}_{k'},
\end{align}
for $k, k' \in [K], ~k \neq k'$. Therefore,  $\bm{V}_k$'s are independent and isotropic random matrices. Note that the power constraints in \eqref{eq:Associate} give a lower bound on the power budget, i.e., $P_{\rm max}\geq \max_{k} \sigma_{v,k} /N_t $.

\subsubsection{Uncorrelated Fading Channels}
For massive MIMO systems, i.e., $N_r, N_t \rightarrow \infty$, the high-dimensional distributions of matrices $\bm{H}_{k}^{\mathsf{H}}\bm{H}_k$ asymptotically follow the Marchenko-Pastur law distribution~\cite{marchenko1967distribution}, thus leading to  Bai-Yin laws~\cite{bai2008limit}, \cite{vershynin2010introduction} for the extreme eigenvalue of the matrix $\bm{H}_{k}^{\mathsf{H}}\bm{H}_k$ with Wishart distribution. Specifically, as the dimensions $N_r, N_t$ increase to infinity while the aspect ratio ${N_t}/{N_r}$ is kept fixed, the following relations between the maximum and minimum eigenvalues of the matrix $\bm{H}_{k}^{\mathsf{H}}\bm{H}_k$ hold \cite{vershynin2010introduction,bai2008limit}:
\begin{subequations}
     \begin{align}
    \lambda_{\rm min}(\bm{H}_{k}^{\mathsf{H}}\bm{H}_k)   & \approx  (N_r - \sqrt{N_t})\sigma_{h,k}^2, \\   \lambda_{\rm max}(\bm{H}_{k}^{\mathsf{H}}\bm{H}_k)  \label{eq:Machenko}
      & \approx (N_r+ \sqrt{N_t})\sigma_{h,k}^2, 
 \end{align}
\end{subequations}
 for $k \in [K]$. Moreover, the eigenvalues of Gaussian matrices $\bm{H}_k$ based on Wigner’s semicircle law yield the following results on the largest eigenvalues, also known as Tracy-Widom law~\cite{feldheim2010universality}: 
 \begin{align}
    \label{eq:Tracy}
     \lambda_{\rm max}(\bm{H}_{k})  \rightarrow (\sqrt{N_r}+ \sqrt{N_t})\sigma_{h,k}^2, \quad k \in [K]. 
 \end{align}
 Note that when ${N_t}/{N_r} \rightarrow 0$ for a relatively large number of antennas at the CP, i.e.,  $N_r \gg 1$,  we obtain the following properties~\cite{marzetta2010noncooperative}:
\begin{subequations}
\label{eq:Massive_total}
\begin{align}
    \label{eq:Massive1}
    \frac{\bm{H}_{k'}^{\mathsf{H}}\bm{H}_k}{N_r} & \approx \bm{0}, \forall~k, k' \in [K], ~k \neq k', \\
    \label{eq:Massive2}
    \frac{\bm{H}_{k}^{\mathsf{H}}\bm{H}_k}{N_r} & \approx \sigma_{h,k}^2\bm{I}_{N_t}, \forall~k \in [K],\\
    \frac{\bm{H}_{k}^{\mathsf{H}}\bm{z}}{N_r} & \approx \bm{0}, \forall~k \in [K],
\end{align}
\end{subequations}
where $\sigma_{h,k}^2$ denotes the path loss from node $k$ to the \ac{CP}. Note that the large-scale fading coefficients are assumed to be the same for different antennas at the same node but are node-dependent~\cite{lu2014overview}.  Therefore, we have the following remark. 
\begin{remark}
    In a high dimensional regime where transmitters and the receiver possess a large number of antennas, i.e., $N_r, N_t \gg 1$,~\eqref{eq:Machenko} and~\eqref{eq:Tracy} state that the eigenvalues of matrices $\bm{H}_k^{\mathsf{H}}\bm{H}_k/N_r$ and $\bm{H}_{k'}^{\mathsf{H}}\bm{H}_k$ concentrate around $(1-\sqrt{N_t}/N_r)\sigma_{h,k}$ and zero, respectively for $k=k'$ and $ k\neq k', k, k'\in [K]$. Therefore,  by setting $\sqrt{N_t}/N_r \approx 0$, we can asymptotically reduce the fluctuations around the expected values $(1-\sqrt{N_t}/N_r)\sigma_{h,k}$, which removes the effects of small scale fading from the wireless channels.  
\end{remark}   

By substituting~\eqref{eq:Massive_total} into~\eqref{eq:MIMO-setup-2}, the term $\bm{y}_{\rm inter}$ asymptotically goes to zero and we obtain 
\begin{align}
    \label{eq:MIMO-setup-4}
      \bm{y} \approx  \frac{N_r}{\beta} \sum\nolimits_{k=1}^K \sigma_{h,k}^2\bm{V}_k^{\mathsf{H}}\bm{V}_k\bm{x}_k. 
\end{align}
Using the isotropy property of beamforming matrices, then $\bm{V}_k^{\mathsf{H}}\bm{V}_k \approx \sigma_{v,k}^2 N_t\bm{I}_{L}$. Also, to compensate for the large-scale fading, we set $\sigma_{v,k}^2 = 1/ \sigma_{h,k}^2$ for all $k\in [K]$. Hence, we denote the approximate received symbols as
\begin{align}
    \label{eq:MIMO-setup-5}
      \hat{\bm{r}}  \approx \sum\nolimits_{k=1}^K \bm{x}_k.
\end{align}
Hence, we can recover the sum signal $\sum\nolimits_{k=1}^K \bm{x}_k$ over the fading channel.

\begin{remark}
We note that the proposed beamforming scheme performs well as long as the channel matrices for all nodes are statistically independent, without requiring specific assumptions regarding the distributions of the beamforming matrices. However, isotropic and independent properties help reduce the variance of the error term rapidly. Also, we note that our CSI-unaware fading compensation targets \emph{small-scale} (fast) fading, which is hard to estimate and varies rapidly. \emph{Large-scale} factors (pathloss/shadowing) are assumed to be known at the transmitters and handled via power control.
\end{remark}

\subsubsection{Correlated Fading Channels}
Due to the isotropic and independent properties of the distribution of the generated $\bm{V}_k$'s, the interference term $\bm{y}_{\rm inter}$ remains zero even in the case that users' channels are correlated, i.e.,  $\bm{H}_{k'}^{\mathsf{H}}\bm{H}_k \neq 0$ for all $k, k' \in [K]$. Indeed, in the case where the channel of different nodes are correlated, the massive MIMO system gives us  $\bm{H}_{k'}^{\mathsf{H}}\bm{H}_k \approx N_r \alpha_{k,k'} \bm{I}_{N_t}$ where $\alpha_{k,k'}$ is the correlation factor between the channels for users $k$ and $k'$ for $k\neq k'$, and $(k, k')\in [K^2]$. Hence,   $\bm{y}_{\rm inter}$ becomes
\begin{align}
    \label{eq:inter-approx}
      \bm{y}_{\rm inter} = N_r \alpha_{k,k'}  \sum\nolimits_{k,k',k\neq k' }^K\bm{V}_{k'}^{\mathsf{H}} \bm{V}_k\bm{x}_k.
\end{align}
Then, invoking the independency property of distribution $\mathcal{F}_k$ from~\eqref{eq:Independency}, we have $\bm{V}_{k'}^{\mathsf{H}} \bm{V}_k \approx \bm{0}$, which leads to 
\begin{align}
    \label{eq:inter-approx-2}
      \bm{y}_{\rm inter}  \approx N_r \alpha_{k,k'}  \bm{0} \times \sum\nolimits_{k=1}^K\bm{x}_k = \bm{0}.
\end{align}
Therefore, employing a large number of antennas at the receiver and transmitter asymptotically diminishes the channel effects.

\begin{remark}
    During the fading compensation, only the \ac{CP} needs to access the \ac{CSI} for generating $\bm{U}$ from $\bm{H}_k$ and $\bm{V}_k$'s. Nodes do not need to communicate their beamforming matrix $\bm{V}_k$ every round. Instead, they can share their random seed for generating the matrix $\bm{V}_k$ from the distribution $\mathcal{F}_k$ with the \ac{CP} to avoid the communication overhead for large metrics $\bm{V}_k$'s. In this case, every node only sends a scalar to \ac{CP} at the beginning of the computation procedure. We emphasize that nodes must generate $\bm{V}_k$ i.i.d. according to a specific distribution to ensure statistical fading compensation. 
\end{remark}

\begin{remark}
     Within the massive MIMO framework, averaging small-scale fading effects yields a more stable sum-channel estimation, rendering it less prone to errors. Specifically, if $\bm{H}_{k} \sim \mathcal{CN}\left(\bm{0}, \sigma_{h,k}^2\bm{I}_{N_r}\right)$
and, under the isotropy condition stated in \eqref{eq:Isotropy}, $\bm{U} \sim \mathcal{CN}(\bm{0}, {\sum_k \sigma_{v,k}^2 \sigma_{h,k}^2}/{N_r^2}\bm{I}_{N_r}),$
where \(\sigma_{v,k}^2\) represents the variance of \(\bm{V}_k\). Furthermore, because the nodes can compensate for large-scale fading, the distribution simplifies to $\bm{U}\sim \mathcal{CN}(\bm{0}, {K}/{N_r^2}\bm{I}_{N_r}).$ Consequently, the receiver beamforming matrix approximates an identity matrix, thereby simplifying its estimation.
\end{remark}

Nonetheless, an open question remains: what is the minimum number of receiver antennas required to achieve a desired level of fading error performance? In the next section, we conduct a non-asymptotic analysis to elucidate the relationship between the receiver antenna count and the fading error.

\subsection{How Well Can We Compensate for the Channel Effect?}

The previous section discussed the asymptotic elimination of channel effects, including fading and noise, for a sufficiently large number of receiver antennas $N_r$. Despite this, practical constraints exist on the maximum number of antennas that can be implemented. While including additional antennas improves system capacity, it simultaneously introduces greater complexity. Together with the industry, we must determine an optimal antenna count that adjusts the advantages of increased capacity with the associated complexity. 

Therefore, an in-depth examination of how the system capacity fluctuates with the increment in antennas in massive MIMO systems is needed. To this end, we propose a non-asymptotic analysis of the computation error. In the following theorem, we establish a probabilistic upper bound on the MSE.

{\change
\begin{theorem}\label{th:NError}
Assume a communication network with $K$ nodes equipped with $N_t$ transmitter antennas and the \ac{CP} with $N_r$ receiver antennas. Let us define the aggregated signal $\bm{r}:= \sum_{k=1}^K\bm{x}_k \in \mathbb{C}^{L}$, which is generated by the transmitted signal of node $k$, $\bm{x}_k$, over the fading channel with coefficients $\bm{H}_k\sim \mathcal{CN}(\bm{0}, N_r\sigma_{h,k}^2\bm{I}_{N_t})$, and beamforming matrices $\bm{V}_k$, which satisfy the isotropy and independency properties from~\eqref{eq:Isotropy} and~\eqref{eq:Independency}, respectively,  for $k\in [K]$. Also, let us denote by $\bm{y}$ the received signal at the \ac{CP} as in~\eqref{eq:MIMO-setup-4}. Then, the error norm between the estimated value of $\hat{\bm{r}}:= \bm{y}/\beta$ and $\bm{r}$  is upper bounded by $\epsilon$, i.e.,  
\begin{align}
    \label{eq:epsilonupp}
    \|\bm{r}- \hat{\bm{r}} \|_2 & \leq \epsilon, 
\end{align}
in which  $N_t \geq L$ and $N_r$ fulfills the lower bound 
\begin{align}
    \label{eq:Nrnumber}
    N_r & \geq  \max\Big\{  \frac{2c_{\sigma, \gamma} LK^2\gamma \sigma_z^2}{\epsilon^{2}}\ln{\Big(\frac{2K(L+1)}{\delta}\Big)}, L\Big\}, 
\end{align}
with probability no less than $1-\delta$, where $\gamma={\sum_{k=1}^K\|\bm{x}_k\|_2^2}$  represent the transmitter  signal power and $c_{\sigma, \gamma}:= \min\{{\sigma_z^2}, \gamma^2 \}/16$.
\end{theorem}
\begin{proof}
    The proof is provided in Appendix~\ref{sec:ProofTheorem}.
\end{proof}
}

The important takeaway from Theorem~\ref{th:NError} is that the number of receiver antennas, $N_r$, has an inverse relation concerning the computation error variance, $\epsilon^2$, i.e., $N_r = \mathcal{O}(\epsilon^{-2})$. {\change Additionally, the error variance, $\epsilon^2$, remains unchanged irrespective of the number of nodes, $K$, in the network. However, maintaining the same error level as the network size increases requires the number of receiver antennas to grow quadratically with $K$. This quadratic scaling originates from the statistical dependence among the aggregated interference terms in \eqref{eq:signalinter}, which limits the effectiveness of averaging and consequently slows down the convergence of the CSI compensation scheme.

We note that the tolerance $\epsilon$ is application-driven. We first need to choose $\epsilon$ based on the application, (e.g., distributed learning, sensing networks)  and then set $(N_{r},N_{t})$ according to Theorem \ref{th:NError}.
}



\begin{remark}
    Note that the provided analysis in Theorem~\ref{th:NError} can be extended for any sub-Gaussian channel noise $\bm{z}$.  In the case of heavy-tailed distributions for the channel noise, to mitigate the influence of outliers, we need to consider more advanced techniques, such as M-estimator~\cite{lucas1997robustness}, a Bayesian framework for designing beamforming matrices~\cite{bell2000bayesian}. 
\end{remark}

While Theorem~\ref{th:NError} guarantees $\|\bm{r}- \hat{\bm{r}} \|_2\le\epsilon$ with $N_r=\mathcal{O}(\epsilon^{-2})$, performance degrades when $N_r$ is small—an inherent cost of our CSI-unaware transmitter design. In practice, comparable error targets can be approached with modest arrays by (i) reducing the number of functions $L$, (ii) averaging across additional time/frequency rounds, and (iii) employing stronger channel coding. If transmitter-side CSI were available (as in CSI-aware OAC\cite{zhu2018mimo}), pre-equalization could compensate channel effects and mitigate small-array loss.

 {\change
 
 \subsection{Imperfect CSI at the Receiver}

The proposed framework assumes perfect CSI for beamforming design at the CP.  In practice, imperfections in CSI manifest as an additional distortion term in the aggregated signal observed at the CP, which degrades the computation accuracy.
To characterize the sensitivity of the proposed scheme to CSI uncertainty, we model the imperfect CSI using channel estimation error model from~\cite{wang2007performance}. In particular, for node $k$, the estimated channel matrix is modeled as
\begin{align}
    \hat{\bm{H}}_k = \bm{H}_k + \Delta\bm{H}_k,
\end{align}
where $\bm{H}_k$ denotes the true channel realization and $\Delta\bm{H}_k$ represents the estimation error, assumed to be statistically independent of $\bm{H}_k$.  The estimation error is usually modeled as $\Delta\bm{H}_k$ , whose i.i.d. entries distributed as $\mathcal{CN}({0}, \sigma_{I, k}^2)$.  Given this model, the CP selects $\bm{U} = \sum_{k=1}^K \bm{V}_k^{\mathsf{H}} \hat{\bm{H}}_k^{\mathsf{H}} / N_r$. Under imperfect CSI, the received signal can be decomposed as $\tilde{\bm{y}} = \bm{y} + \Delta\bm{y}$, where $\bm{y}$ corresponds to the perfect CSI aggregation in \eqref{eq:MIMO-setup-2} and $\Delta\bm{y}$ captures the distortion induced by channel estimation errors. This distortion term is given by
\begin{align}
    \hspace{-5pt}\Delta\bm{y} 
     =  \frac{1}{\beta}\sum_{k,k'=1}^K 
    \bm{V}_{k'}^{\mathsf{H}} \Delta\bm{H}_{k'}^{\mathsf{H}} \bm{H}_k \bm{V}_k \bm{x}_k  +\frac{1}{\beta} \sum_{k=1}^K \bm{V}_k^{\mathsf{H}} \Delta\bm{H}_k^{\mathsf{H}} \bm{z}.
    \label{eq:imperfect_csi_distortion}
\end{align}
Due to the statistical independence between all $\bm{H}_k$, $\Delta\bm{H}_k$, and the noise vector $\bm{z}$ \cite{nosrat2011mimo}, and leveraging the channel hardening property, the distortion term vanishes asymptotically as the number of receive antennas grows. In particular, we have $\lim_{N_r \to \infty} \Delta\bm{y} = \bm{0},$ which consequently give us $\lim_{N_r \to \infty} \tilde{\bm{y}} = \bm{y} = \sum_{k=1}^K \bm{x}_k.$ Hence, imperfect CSI does not induce a persistent bias in the large-antenna regime. In other words, $\Delta\bm{y}$ effectively behaves as an additional interference or noise component. Defining $\tilde{\bm{z}} := \bm{z} + \Delta\bm{y}$, its variance $\sigma_{\tilde{z}}^2$ scales as $\sigma_z^2 +  \sum_k\sigma_{I, k}^2$. Incorporating this effect leads to the following Theorem .

\begin{theorem}\label{cor:new-imperfct}
Under the same conditions as Theorem~\ref{th:NError}, and considering imperfect CSI at the receiver modeled by $\Delta\bm{H}_k$ whose i.i.d. entries distributed as  $\mathcal{CN}({0}, \sigma_{I, k}^2)$ for $k\in [K]$, the computation error bound in \eqref{eq:epsilonupp} holds provided that $N_r, N_t\geq L$ and $N_r$ satisfies
\begin{align} \label{eq:Nrnumbercor-2} 
    N_r \geq  8\frac{c_{\sigma, \gamma} LK^2\gamma \sigma_z^2 \max\{\bar{\xi}, 1\}}{\epsilon'^{2}}\ln{\Big(\frac{4K(L+1)}{\delta}\Big)}. 
\end{align}
with probability at least $1-\delta$, where  $\bar{\xi} = \sum_{k}\sigma_{I, k}^2\sigma_{h,k}^{-2}/K$.
\end{theorem}
\begin{proof}
    The proof is provided in Appendix~\ref{sec:new-imperfct}. 
\end{proof}

Theorem~\ref{cor:new-imperfct} indicates that, as long as $\bar{\xi}\leq 1$,
the presence of channel estimation errors does not increase the required
number of receive antennas compared with the perfect CSI case.
Equivalently, no additional antennas are needed to achieve the same aggregation accuracy. The condition $\bar{\xi} = 1$ corresponds to a normalized estimation error power on the order of the channel power, i.e., $\sigma_{I,k}^2/\sigma_{h,k}^2\approx 1$. In practical systems, channel estimation errors are typically well below this level (often within approximately $20\%$~\cite{wang2007performance}), which indicates that the proposed VeComp scheme
is robust to CSI imperfections at the receiver.

 }

\section{VecComp for Multiple Function Computation}\label{sec:FunctionComp}

So far, we have designed beamforming matrices to compensate for the effect of the channel in performing function computation. In this section, we introduce the VecComp approach for computing matrix functions by designing the encoders $\mathscr{E}_k$s and decoder $\mathscr{T}$. VecComp has two distinct setups: 

\begin{itemize}
    \item \textbf{Exact Setup}:  the channel effects are entirely compensated through the beamforming technique outlined in the previous section.
    \item \textbf{Inexact Setup}: the received signal is contaminated with a certain error tolerance, e.g., $\eta >0$.
\end{itemize}

\begin{figure}
    \centering

\scalebox{0.75}{

\tikzset{every picture/.style={line width=0.75pt}} 
\begin{tikzpicture}[x=0.75pt,y=0.75pt,yscale=-0.65,xscale=0.65]

\draw    (390,180) -- (610,180) ;
\draw    (500,289.98) -- (500,68.22) ;

\begin{scope}[shift={(0cm,6cm)}]
\draw (100, 0) node  {\footnotesize $\Re(x_2)$};
\draw (220, 100) node  {\footnotesize $\Im(x_2)$};
\draw (160, 80) node  {\small$1$};
\draw (35,80) node {\small$-1$};
\draw (112,170) node  {\small$-j$};
\draw (110,30) node  {\small $j$};

\draw    (15,100) -- (185,100) ;
\draw    (100,180) -- (100,20) ;

\draw [dash pattern={on 4.5pt off 4.5pt}] (101,100) node{}  circle  (51);
\draw [fill=black!] (101,50) node{}  circle  (3);
\draw [fill=black!] (101,150) node{}  circle  (3);
\draw [fill=black!] (50,100) node{}  circle  (3);
\draw [fill=black!] (151,100) node{}  circle  (3);

\end{scope}

\begin{scope}[shift={(0cm,0cm)}]
\draw (100, 0) node  {\footnotesize $\Re(x_1)$};
\draw (220, 100) node  {\footnotesize $\Im(x_1)$};
\draw (160, 80) node  {\small$1$};
\draw (35,80) node {\small$-1$};
\draw (112,170) node  {\small$-j$};
\draw (110,30) node  {\small $j$};

\draw    (15,100) -- (185,100) ;
\draw    (100,180) -- (100,20) ;

\draw [dash pattern={on 4.5pt off 4.5pt}] (101,100) node{}  circle  (51);
\draw [fill=black!] (101,50) node{}  circle  (3);
\draw [fill=black!] (101,150) node{}  circle  (3);
\draw [fill=black!] (50,100) node{}  circle  (3);
\draw [fill=black!] (151,100) node{}  circle  (3);

\end{scope}

\draw [color=black, fill=black] (400,180) node{}  circle  (3);
\draw [color=black, fill=black] (600,180) node{}  circle  (3);
\draw [color=chestnut, fill=chestnut] (500,180) node{}  circle  (3);
\draw [color=black, fill=black] (500,80) node{}  circle  (3);
\draw [color=black, fill=black] (500,280) node{}  circle  (3);
\draw [color=black, fill=black] (440,240) node{}  circle  (3);
\draw [color=black, fill=black] (560,240) node{}  circle  (3);
\draw [color=black, fill=black] (440,120) node{}  circle  (3);
\draw [color=black, fill=black] (560,120) node{}  circle  (3);


\draw  [dash pattern={on 0.84pt off 2.51pt}] (565.47,204.58) -- (528.75,243.99) -- (475.43,245.39) -- (436.76,207.96) -- (435.38,153.63) -- (472.11,114.22) -- (525.42,112.82) -- (564.09,150.25) -- cycle ;
\draw  [dash pattern={on 0.84pt off 2.51pt}]  (564.09,150.24) -- (615.42,136.06) ;
\draw  [dash pattern={on 0.84pt off 2.51pt}]  (525.42,112.81) -- (544.11,60) ;
\draw  [dash pattern={on 0.84pt off 2.51pt}]  (472.11,114.22) -- (448.76,63.38) ;
\draw  [dash pattern={on 0.84pt off 2.51pt}]  (435.38,153.63) -- (383.25,141.97) ;
\draw  [dash pattern={on 0.84pt off 2.51pt}]  (460.36,300) -- (475.43,245.39) ;
\draw  [dash pattern={on 0.84pt off 2.51pt}]  (547.43,295.77) -- (528.75,243.99) ;
\draw  [dash pattern={on 0.84pt off 2.51pt}]  (616.25,220.56) -- (565.47,204.58) ;
\draw  [dash pattern={on 0.84pt off 2.51pt}]  (384.08,223.1) -- (436.76,207.97) ;


\draw (280,180) node  {\Large $\bigoplus$};

\draw[-latex]    (188.25,135) -- (261.28,168.74) ;

\draw[-latex]    (196.25,242)  -- (258.63,193.83) ;

\draw [-latex]   (300,180) -- (345,180) ;

\draw (250,137) node [anchor=north west][inner sep=0.75pt]  [font=\footnotesize] [align=left] {Over-the-Air};

\draw (440,9.4) node [anchor=north west][inner sep=0.75pt]  [color={rgb, 255:red, 8; green, 80; blue, 177 }  ,opacity=1 ]  {$\mathscr{T}\{{\color{black}r}\} =s_{1} s_{2}$};
\draw (619.76,168.6) node [anchor=north west][inner sep=0.75pt]  [color={rgb, 255:red, 8; green, 80; blue, 177 }  ,opacity=1 ] [align=left] {$0$};
\draw (565.39,250.55) node [anchor=north west][inner sep=0.75pt]  [color={rgb, 255:red, 8; green, 80; blue, 177 }]  [align=left] {$0$};
\draw (558.39,90.55) node [anchor=north west][inner sep=0.75pt] [color={rgb, 255:red, 8; green, 80; blue, 177 }]  [align=left] {$0$};
\draw (505,160) node [anchor=north west][inner sep=0.75pt] [color=chestnut]  [align=left] {$0$};
\draw (493.39,42.55) node [anchor=north west][inner sep=0.75pt]  [color={rgb, 255:red, 8; green, 80; blue, 177 }] [align=left] {$1$};
\draw (420.39,91.55) node [anchor=north west][inner sep=0.75pt]  [color={rgb, 255:red, 8; green, 80; blue, 177 }]  [align=left] {$3$};
\draw (370.39,167.55) node [anchor=north west][inner sep=0.75pt]  [color={rgb, 255:red, 8; green, 80; blue, 177 } ] [align=left] {$9$};
\draw (416.39,238.55) node [anchor=north west][inner sep=0.75pt]  [color={rgb, 255:red, 8; green, 80; blue, 177 }] [align=left] {$ 6$};
\draw (494.39,300.55) node [anchor=north west][inner sep=0.75pt]  [color={rgb, 255:red, 8; green, 80; blue, 177 }  ,opacity=1 ] [align=left] {$4$};
\draw (480,160) node [anchor=north west][inner sep=0.75pt]  [color=chestnut]  [align=left] {$2$};

\end{tikzpicture}}
    \caption{The overlaps of the reshaped constellation points of QPSK modulation do not allow us to compute the product function. }
    \label{fig:ProductQAM}
\end{figure}
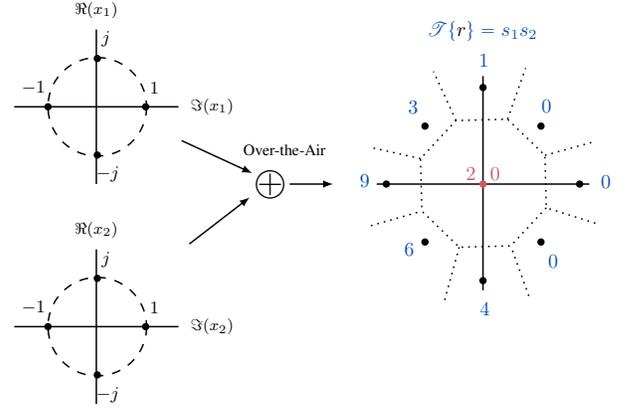

\subsection{VecComp: Function Computation in Exact Setup}

In the exact setup, we assume that the received signal by the \ac{CP} is error-free, i.e., $\bm{r}:= \sum_{k=1}^K\bm{x}_k$. The \ac{CP} applies the tabular function $\mathscr{T} : \mathbb{C}^{L} \mapsto \mathbb{R}^{L}$ on the induced  signal $\bm{r}$ to compute the desired function $\bm{f}(\bm{s}) \in \mathbb{R}^{L}$ function, i.e.,
\begin{align}
    \mathscr{T}\Big(\sum\nolimits_{k=1}^K\mathscr{E}_k({s}_k)\Big) \approx \bm{f}(\bm{s}),
\end{align}
where $\mathscr{E}_k(s_k): = [\mathscr{E}_{k,1}(s_k), \ldots, \mathscr{E}_{k,L}(s_k)]^{\mathsf{T}}$. Because  the encoding operator $\mathscr{E}_k(s_k)$ acts element-wise on the input scalar $s_k$, we decompose the tabular function $\mathscr{T}$ into a sets of $L$ different maps as $\mathscr{T}(\bm{r}) =  \bigoplus_{\ell=1}^L\mathscr{T}_{\ell}(r_{\ell})$, in which, every $ \mathscr{T}_{\ell}: \mathbb{C} \mapsto \mathbb{R}$ maps the element $\ell$ of received vector $y_{\ell}$ to output $\ell$ of function $\bm{f}(\bm{s})$ as
\begin{align}
\label{eq:elem-comp}
    \mathscr{T}_{\ell}\Big(\sum\nolimits_{k=1}^K\mathscr{E}_{k,\ell}(s_k)\Big) = f_{\ell}(s_1, \ldots, s_K).  
\end{align}
We must guarantee that the employed encoders $\mathscr{E}_{k,\ell}$ and tabular functions $\mathscr{T}_{\ell}$  for nodes can accurately compute the desired function to hold the equality in~\eqref{eq:elem-comp}. Figure~\ref{fig:ProductQAM} shows a simple case for $K=2$ nodes that want to compute the product function $f(s_1,s_2) = s_1s_2$ with $L=1$. The constellation points shaped by the nodes cannot be uniquely mapped to the product function.  Indeed, the induced constellation diagram of $\sum_{k=1}^K\mathscr{E}_k({s}_k)$ must cover all the points in the range of function $\bm{f}(\bm{s})$~\cite{saeed2023ChannelComp}. In other words, for any distinct value in the range of function $\bm{f}(\bm{s})$, there must exist a corresponding distinct constellation point in the diagram of $\sum_{k=1}^K\mathscr{E}_k({s}_k)$. This condition can be expressed in the following proposition. 

\begin{prop}
 \label{prop:Vecfeas}
For $\ell \in [L]$, let $f_{\ell}^{(i)}$  be the $i$-th output of the function $f_{\ell}$ for given  values of input ${s}_1^{(i)},{s}_2^{(i)},\ldots,{s}_k^{(i)} \in \mathbb{F}_{Q}^K$ from $ i \in [M]$, where $M$ denotes the number of possible values of the function ${f}_{\ell}$. Further, ${r}_{\ell}^{(i)} \in \mathbb{C}$ represents the corresponding constellation points to ${f}_{\ell}^{(i)}$, i.e., ${r}_{\ell}^{(i)}=\sum_{k=1}^K\mathscr{E}_{\ell, k}({s}_k^{(i)})$. Then, using  the set of encoders $\mathscr{E}_{k,\ell}(\cdot)$ and the tabular functions $\mathscr{T}_{\ell}(\cdot)$ for $\ell \in [L]$,  the computation in~\eqref{eq:elem-comp} is correctly  performed over the \ac{MAC} if and only if we have the following: 
\begin{align}
    {\rm if} & ~~ {f}_{\ell}^{(i)} \neq {f}_{\ell}^{(j)} ~~\text{then}~~{r}_{\ell}^{(i)} \neq {r}_{\ell}^{(j)}, \quad \forall (i,j)\in [M]^{2}.
\end{align}
Given feasibility of the computation over the \ac{MAC}, there exists a map $\mathscr{T} = \bigoplus_{\ell=1}^L\mathscr{T}_{\ell}$ such that $\mathscr{T}\Big(\sum_{k=1}^K\mathscr{E}_k({s}_k)\Big)  = \bm{f}(\bm{s})$, i.e., the desired function $f$ can be correctly computed.
\end{prop}
\begin{proof}
    Let us consider ${f}_{\ell}^{(i)} = {f}_{\ell}^{(j)} = d_{\ell}$. Then, regardless of the values ${r}_{\ell}^{(i)}$ and ${r}_{\ell}^{(j)}$, we  can define a tabular map $\mathscr{T}_{\ell}$ such that
    $\mathscr{T}_{\ell}({r}_{\ell}^{(i)}) = \mathscr{T}_{\ell}({r}_{\ell}^{(i)}) = d_{\ell}$. Therefore, this shows that we need distinct constellation points only for distinct output values of the function $f_{\ell}$ as stated in the theorem.  
\end{proof}

The induced diagram of the  constellation point $i$ at antenna $\ell$ in terms of the transmitted modulation vectors of the nodes can be expressed as
\begin{align}
    \label{eq:constpoint}
    {r}_{\ell}^{(i)} = \langle \bm{a}_{\ell,i}, \bm{X}_{\ell} \rangle,
\end{align}
where $\bm{X}_{\ell}:= [\bm{X}_{\ell,1}, \ldots, \bm{X}_{\ell,K} ]^{\mathsf{T}}\in \mathbb{C}^{QK}$ whose $[\bm{X}_{\ell,k}]_{q} = x_{k,\ell}^{q}$ denotes $q$-th elements of the constellation points owned by node $k$ at antenna $\ell$, for $(q,k)\in [Q]\times [K]$. Also, $\bm{a}_{\ell,i}$ is a binary vector whose elements are determined such that it selects all the constellation points corresponding to function value $f_{\ell}^{(i)}$.

Based on the Proposition~\ref{prop:Vecfeas},  to have a valid computation over the \ac{MAC} for the desired function $\bm{f}(\bm{s}): \mathbb{F}_{Q}^{K} \mapsto \mathbb{R}^L$, we pose the following optimization problem 
\begin{align}
    \nonumber
 \mathcal{P}_{\ell}^{(1)}  = & {\rm find} ~~~\bm{X}_{\ell}~~ \\ \label{eq:feasibility}
& {\rm s.t.}~~
{r}_{\ell}^{(i)} \neq {r}_{\ell}^{(j)},~ 
 \forall (i,j) \in \Omega,~ \|\bm{X}_{\ell}\|_{\rm F}^2 \leq 1,
\end{align}
where  $\Omega \subseteq [M^2]$ whose every element $(i,j)$ is selected as ${f}_{\ell}^{(i)}\neq {f}_{\ell}^{(j)}$, and   $P$ is the allocated power to all node's modulation vectors. For the constraints in~\eqref{eq:feasibility} on valid computation,  we use the given notation  in~\eqref{eq:constpoint} as 
\begin{align}
    \nonumber
   \langle \bm{a}_{\ell,i},\bm{X}_{\ell} \rangle & \neq \langle \bm{a}_{\ell,j},\bm{X}_{\ell} \rangle, \\ \nonumber
    \langle \bm{a}_{\ell,i} -  \bm{a}_{\ell,j}, \bm{X}_{\ell} \rangle  & \neq 0, \\
    \langle \bm{\alpha}_{\ell}^{i,j}, \bm{X}_{\ell} \rangle  & \neq 0, \quad \forall (i,j)\in \Omega.
\end{align}
The last not equal term can be equivalently written as $\bm{X}_{\ell} \notperp \bm{\alpha}_{\ell}^{i,j} $   for  $ (i,j)\in \Omega$.  Let us define set $\mathcal{A}^{\Omega}_{\ell} : = \{\bm{\alpha}_{\ell}^{i,j}, (i,j)\in \Omega\}$, which involves all vectors $\bm{\alpha}_{\ell}^{i,j}$. Then, Problem  $\mathcal{P}_{\ell}^{\rm exact}$ can be equivalently written as
\begin{align}
   \nonumber
 \mathcal{P}_{\ell}^{(1)} = &  {\rm find} ~~~\bm{X}_{\ell}~~ \\
  \label{eq:feasibility1}
& {\rm s.t.}~~  \langle \tilde{\bm{\alpha}}_{\ell}, \bm{X}_{\ell} \rangle \neq 0, \forall~\tilde{\bm{\alpha}}_{\ell} \in \mathcal{A}_{\ell}^{\Omega},  ~ \|\bm{X}_{\ell}\|_{\rm F}^2 \leq 1. 
\end{align}
Problem $\mathcal{P}_{\ell}^{\rm exact}$ is a feasibility problem to find a point that satisfies the constraints, and the solution is not unique. Indeed, $\mathcal{P}_{\ell}^{\rm exact}$ finds a vector not perpendicular to a given set of vectors in $\mathcal{A}^{\Omega}_{\ell}$. One possible way to obtain a solution for Problem $\mathcal{P}_{\ell}^{\rm exact}$ is given by Algorithm~\ref{Alg:non-orthogonal}. The following proposition guarantees the feasibility of Algorithm~\ref{Alg:non-orthogonal}.

\begin{algorithm}[!t]
\caption{Find non-orthogonal vector}\label{Alg:non-orthogonal}
\begin{algorithmic}[1]
\Procedure{\rm NonOrth}{$\bm{v}_1, \ldots, \bm{v}_m$}
\State Reorder vectors such that \( \{\bm{v}_2, \ldots, \bm{v}_k\} \perp \bm{v}_1  \) are orthogonal and  \( \{\bm{v}_{k+1}, \ldots, \bm{v}_m\} \notperp \bm{v}_1  \) are not
\State Recursively, $ \bm{y} : = {\rm NonOrth}(\bm{v}_2, \ldots, \bm{v}_k)$
\State $\alpha = \max \left\{ \left| \frac{\langle \bm{y}, \bm{v}_i \rangle}{\langle \bm{v}_1, \bm{v}_i \rangle} \right| : i \in \{k+1, \ldots, m\} \right\} + 1.$
\State \( \bm{x} \leftarrow \alpha \bm{v}_1 + \bm{y} \)

\Return \( x \)
\EndProcedure
\end{algorithmic}
\end{algorithm}

\begin{prop}\label{prop:SolP1}
    Let $\hat{\bm{Y}}_{\ell} = {\rm NonOrth}(\mathcal{A}^{\Omega}_{\ell}) \in \mathbb{C}^{QK}$ be the output from Algorithm~\ref{Alg:non-orthogonal}. Then, $\hat{\bm{X}}_{\ell} =(\hat{\bm{Y}}_{\ell} + \tilde{\bm{X}}_{\ell})/\kappa$, where $\kappa = \sqrt{\|\hat{\bm{Y}}_{\ell}\|_2^2 +\|\tilde{\bm{X}}_{\ell}\|_2^2}$ and $\tilde{\bm{X}}_{\ell} \in \mathcal{N}(\mathcal{A}^{\Omega}_{\ell})$, is a solution to Problem~$\mathcal{P}_{\ell}^{\rm exact}$.
\end{prop}
\begin{proof}
    See Appendix~\ref{sec:ProofP1prob}. 
\end{proof}
Given the proposed solution in Proposition~\ref{prop:SolP1}, we can design the encoders as nonlinear operators that map the input value $\bm{s}$ to the set of the obtained constellation points. In particular, we define set $\mathcal{S}_k$ involving all possible values of $s_k\in \mathbb{F}_{Q}$, and $\mathcal{X}_k$ involves all the values of column $k$ of $\hat{\bm{X}}_{\ell}$   for $\ell \in [L]$ . Then,  $ \mathscr{E}_k$ becomes the bijective function that  maps the elements of $\mathcal{S}_k$ to the elements of set $\mathcal{X}_k$, i.e., $\mathscr{E}_k: \mathcal{S}_k \mapsto \mathcal{X}_k$ and
\begin{align}
    \label{eq:ENcoders}
    \hat{\mathscr{E}}_{k,\ell} = s_k^{(q)} \mapsto \hat{x}_{k,\ell}^{(q)}, \quad \forall (\ell, q) \in [L]\times [Q]. 
\end{align}
Similarly, having access to the encoders, the tabular function  $\mathscr{T}$ can be determined uniquely. Indeed, we define set $\mathcal{X} := \bigoplus_{k=1}^K\mathcal{X}_k$, and define $\mathcal{W}_{\ell}^f$ as the range of function ${f}_{\ell}(\bm{s})$ for $\ell \in [L]$. Then, the tabular function $\mathscr{T}_{\ell}(\cdot): \mathcal{X} \mapsto \mathcal{W}_{\ell}^f$ is determined as the map 
\begin{align}
    \hat{\mathscr{T}}_{\ell} = \sum_{k=1}\hat{x}_{k,\ell}^{(q)} \mapsto {f}_{\ell}({s}_1^{(q)}, \ldots, {s}_K^{(q)}), \quad \forall \ell \in [L]. 
\end{align}
Consequently, we get $\hat{\mathscr{T}}:= \bigoplus_{\ell=1}^L \hat{\mathscr{T}}_{\ell}$.
\begin{remark}
So far, we have provided a mechanism to compute a multivariate function $\bm{f}(\bm{s}) \in \mathbb{R}^{L}$ in a distributed manner for a network consisting of $K$ nodes and the CP server, i.e., $\mathscr{T}\Big(\sum_{k=1}^K\mathscr{E}_k({s}_k)\Big)  = \bm{f}(\bm{s})$. Giving the desired function $\bm{f}(\bm{s})$, the \ac{CP} compute the modulation constellations $\hat{\bm{X}}_{\ell}$ from Proposition~\ref{prop:SolP1}. Then, the CP shares the encoder $\mathscr{E}_k(\cdot)$ to node $k$ for $k\in [K]$. Subsequently, node $k$ build the encoder $\mathscr{E}_{k}$ and employs its modulation vector $\bm{x}_k$ for computing the function $\bm{f}(\bm{s})$.  
\end{remark}
\subsection{VecComp: Function Computation for Inexact Setup}\label{sec:FuncIneaxxt}

Let $\bm{e}$ be the uncompensated  error over the \ac{MAC}, the \ac{CP} receives $\Tilde{\bm{r}} := \bm{r}+ \bm{e}$ where $\|\bm{e}\|_{2} \leq \eta$. For $\eta > 0$, solutions to Problem $ \mathcal{P}_{\ell}^{\rm exact}$ may result in high computation error even for a small value of $\eta$. To ensure the robustness of the solution to the error, we need to modify the constraints for  Problem $ \mathcal{P}_{\ell}^{\rm exact}$ in~\eqref{eq:feasibility1} to capture the computation errors.

Indeed, the angle between $\bm{X}_{\ell}$ and vectors $\bm{\alpha}_{\ell}^{i,j}$ for $ (i,j)\in \Omega$ must be dependent on the value of the corresponding computation error, i.e., $|{f}_{\ell}^{(i)} - {f}_{\ell}^{(j)}|$.  Accordingly, we replace them with a smoother condition, such as: 
\begin{align}
\nonumber
  \mathcal{P}_{\ell}^{(1)} = &  \underset{\mathcal{E} > 0, \bm{X}_{\ell}}{\rm max} ~~~\mathcal{E}~~
 \\ {\rm s.t.}~& 
 |\langle \bm{\alpha}_{\ell}^{i,j}, \bm{X}_{\ell} \rangle|^2 \geq  \mathcal{E}\gamma_{i,j}^{\ell},~~ \|\bm{X}_{\ell}\|_{\rm F}^2 \leq 1, \label{eq:feasibility-non-convex}
\end{align}
for all $(i,j) \in \Omega$, where  $\gamma_{i,j}^{\ell}:=|{f}_{\ell}^{(i)} - {f}_{\ell}^{(j)}|^2 $ penalize the distance between zero with different weight proportional to the computation error $|{f}_{\ell}^{(i)} - {f}_{\ell}^{(j)}|$.  Problem $\mathcal{P}_{\ell}^{(1)}$ is a quadratically constrained quadratic programming (QCQP) problem with non-convex constraints, and it is known to be \textit{NP-hard}~\cite{Sidir2006Physical}.  To circumvent the non-convexity, we use the \textit{lifting trick} by treating $\bm{X}_{\ell}\bm{X}_{\ell}^{\mathsf{T}}$  as matrix $\bm{W}_{\ell}$ for $\ell \in [L]$. Then, following a similar strategy as \cite{saeed2023ChannelComp}, we can reach the following relaxed convex optimization problem, 
\begin{subequations}
    \label{eq:feasibility-convex-lift}
    \begin{align}
\nonumber
  \mathcal{P}_{\ell}^{(2)} & = \underset{\mathcal{E} > 0, \bm{W}_{\ell}}{\rm max}  ~\mathcal{E}~~~~
  {\rm s.t.} ~~ 
 \langle \bm{B}_{\ell}^{i,j}, {\rm trace}(\bm{W}_{\ell}) \rangle  \geq   \mathcal{E}, \\ & ~{\rm trace}(\bm{W}_{\ell}) \leq  1, ~\bm{W}_{\ell} \succeq \bm{0}.
\end{align}
\end{subequations}
The optimization problem denoted as $\mathcal{P}_{\ell}^{(2)}$ is formulated as a semi-definite programming (SDP) problem. By solving this problem, we obtain the optimal weight matrix $\bm{W}_{\ell}^{*}$. Subsequently, the optimal solution  $\bm{X}_{\ell}^{*}$ is computed through the Cholesky factorization of $\bm{W}_{\ell}^{*}$~\cite{luo2010semidefinite}, specifically when $\bm{W}_{\ell}^{*}$ resulted in a rank-one matrix. In instances where $\bm{W}_{\ell}^{*}$ is not a rank-one matrix, a sub-optimal solution for $\mathcal{P}_{\ell}^{(1)}$ is retrievable via Gaussian randomization techniques~\cite{luo2010semidefinite}. The overall procedure is summarized in Algorithm~\ref{alg:Ecnoders}.

 It is worth noting that the modulation design in \eqref{eq:feasibility-convex-lift} extends directly to the $T$-retransmission (time-slot) setting in Remark~3. Concretely, replace $\bm{W}_{\ell}^{*}$ by its best rank-$T$ approximation $\bm{W}_{\ell}^{(T)}$ (e.g., retain the top $T$ eigenvalues/vectors of $\bm{W}_{\ell}^{*}$), and factorize $\bm{W}_{\ell}^{(T)}$ to obtain the $T$-slot encoder. This construction is optimal for the $T$-slot design objective~\cite{yan2025multi} and improves reliability through retransmissions, thereby enabling a controllable reliability–latency tradeoff~\cite{yan2025multi}.

\begin{remark}
 Optimal solutions to each instance of Problem $\mathcal{P}_{\ell}^{(1)}$, where $\ell \in [L]$, enable the computation of the desired function $\bm{f}(\bm{s})$ over the fading \ac{MAC}. Upon attaining $\bm{X}_{\ell}^{*}$, the encoders $\mathscr{E}_k$ are systematically constructed for all $K$ nodes, following the methodology outlined in Eq.~\eqref{eq:ENcoders}. For the decoding sets $\mathscr{T}_{\ell}$, a Voronoi diagram is generated based on the complete set of feasible constellation points, employing a maximum likelihood estimator~\cite{saeed2023ChannelComp,SugiharaVoronoi1992}. Subsequently, decoder $\mathscr{T}_{\ell}$ is defined as a tabular function mapping points within the resultant Voronoi cells to the corresponding output of ${f}_{\ell}(\bm{s})$, for each $\ell \in [L]$.    
\end{remark}

\begin{algorithm}[!t]
\caption{Design the Modulation Vectors}\label{alg:Ecnoders}
\begin{algorithmic}[1]
	\State \textbf{Input:} Function $\bm{f}(s_1,\ldots,s_K)$   
	\State Set $M = Q^K$
	\State \textbf{Output:} Modulation vectors  $\{\bm{X}_{\ell}\}_{\ell=1}^{L}$
	\Procedure{Optimization}{$\{f_{\ell}^{(i)}\}_{(i,\ell)=(1,1)}^{[M]\times [L]} $}
	\For {in parallel $\ell \gets 1,2,\ldots,L$}
	\State Obtain $\bm{W}_{\ell}$ by solving \eqref{eq:feasibility-convex-lift}
    \State  Cholesky decomposition  for $\bm{W}_{\ell} = \bm{L}_{\ell}\bm{L}_{\ell}^{\mathsf{H}}$  
	\State Set $\bm{X}_{\ell} = \bm{L}_{\ell}$
        
	\EndFor
	\EndProcedure
\end{algorithmic}
\end{algorithm}

\begin{remark}\label{rem:complexity}
VecComp raises ChannelComp’s complexity only linearly in the number of functions, $L$. Since ChannelComp’s SDP solver runs in $\mathcal{O}(\max\{n,m\}^4n^{0.5})$~\cite{saeed2023ChannelComp}, VecComp’s overall complexity becomes 
$\mathcal{O}(L\max\{n,m\}^4n^{0.5})$, with decoding scaling as $\mathcal{O}(LQ')$. All other metrics—spectral efficiency and per-transmission energy—remain identical to ChannelComp, with the only extra cost due to standard Massive MIMO multi-antenna reception~\cite{LarssonMassive2014}.
\end{remark}
\begin{remark}\label{rem:nodes}
In VecComp, increasing the number of nodes \(K\) raises both optimization complexity and antenna requirements. For instance, sum-function constraints grow as \(\mathcal{O}(K^2)\) in \eqref{eq:feasibility-convex-lift}, enabling more complex computations at the cost of heavier optimization. Meanwhile, \eqref{eq:Nrnumber} shows the receiver antennas must scale as \(N_r=\mathcal{O}(LK/\epsilon^2)\) to preserve error tolerance \(\epsilon\). Thus, designers must trade off node count against available antennas to balance efficiency and reliability.
\end{remark}

\section{Application Context: Distributed ML and IoT}

VecComp provides a digital modulation framework for computing vector functions directly over the wireless MAC. This capability is particularly relevant in distributed machine learning and IoT systems, where uplink communication constitutes a critical bottleneck~\cite{wen2023task}. Instead of transmitting raw data, each device $k$ encodes its local task vector $\bm{s}_k$—originating from learning or sensing pipelines—into channel symbols. The receiver, equipped with CSI, can then form one-shot linear estimates $f(\bm{x})$ of the desired function while remaining compatible with conventional channel coding, quantization, and MIMO techniques. 

This paradigm reduces uplink traffic, scales efficiently with the number of devices, and naturally supports computational primitives that frequently arise in ML/IoT applications. Typical examples of the desired function include aggregation, affine transformations, and short-window convolutions. To illustrate, the subsequent subsections instantiate VecComp for two representative cases: (i) affine transformations implemented via PAM (Sec.~\ref{sec:PAMspec}), and (ii) convolution operations realized via QAM (Sec.~\ref{sec:QAMSpec}).

\subsection{Special Case I: Affine Transformation with PAM}\label{sec:PAMspec}

One appropriate computation is to use an affine transformation, in which an affine map comprises two functions: a translation and a linear map. This can be useful in applications such as wireless sensor networks and edge computing, in which computations are typically limited to affine transformation (weighted summation)~\cite{amiri2020machine}. Mathematically, an affine transform can be represented by:
\begin{align}
    \bm{y} = \bm{f}(\bm{s}) = \bm{A}\bm{s} + \bm{b}, \quad \bm{s}\in \mathbb{F}_{Q}^K,
\end{align}
where $\bm{A} \in \mathbb{F}_{Q'}^{L\times K}$ represents a linear map with $Q' :=\prod_{k=1}^KQ_{k}$, and $\bm{b} \in \mathbb{F}_{Q}^L$ is the translation function. We can rewrite the transformation as follows:
\begin{align}
\label{eq:affine_function}
    f_{\ell}(s_1,\ldots,s_K) = \sum\nolimits_{k=1}^Ka_{k,\ell}s_k + b_{\ell},\quad \ell \in [L],
\end{align}
where $a_{k,\ell}\in \mathbb{F}_{Q_{K}}$ denotes the entry of matrix $\bm{A}$ at $(\ell,k)$, $b_{\ell}$ is element $\ell$ of vector $\bm{b}$. For ease of explanation, consider the field $\mathbb{F}$ to be an integer field   $\mathbb{Z}$, i.e., $\mathbb{F}_{Q_{k}}= \mathbb{Z}_{Q_{k}}: =\{0,1\ldots,Q_{k}-1\}$ for $k\in [K]$.  Then, the product $a_{k,\ell}s$ belongs to $\mathbb{F}_{QQ_k}$ field, and we can use the following encoder for transmission:
\begin{align}
    \label{eq:Encoder}
    \mathscr{E}_{k,\ell}(s)= a_{k,\ell}s-\Big\lfloor\frac{QQ_{k}}{2}\Big\rfloor,  \quad (\ell, k) \in [L] \times [K], 
\end{align}
in which $\mathscr{E}_{k}$ maps the input value onto the constellation point of a \ac{PAM}. Having encoders $\mathscr{E}_{k}$'s, the decoder accordingly becomes:
\begin{align}
    \mathscr{T}_{\ell}(y)= \mathcal{D}(y) + b_{\ell} +\sum\nolimits_{k=1}^K\Big\lfloor\frac{QQ_{k}}{2}\Big\rfloor,  \quad \ell \in [L],
\end{align}
where
\begin{align}\nonumber
    \mathcal{D}(y) = \begin{cases}
        r,  \quad  r -0.5\leq y\leq r+0.5, \in r\in \mathbb{Z}_{Q_f}, \\
        (1-Q)\sum\nolimits_{k=1}^K(1-Q_k), \quad  y < K(1-Q)+0.5, \\
     \sum\nolimits_{k=1}^K(Q_k-1)(Q-1),  \quad y > \varsigma-0.5,
    \end{cases}
\end{align}
where $\varsigma = \sum_{k=1}^K(Q_k-1)(Q-1)$. Notably,  we observe that the proposed $\mathscr{E}_k$ in~\eqref{eq:Encoder} satisfies all the constraints in~\eqref{eq:feasibility-non-convex}. Therefore, from Proposition~\ref{prop:Vecfeas}, we know that $ \mathscr{T}_{\ell}\Big(\sum_{k=1}^K\mathscr{E}_{k,\ell}(s_k)\Big) = f_{\ell}(s_1, \ldots, s_K)$ holds over ideal \ac{MAC}.

\subsection{Special Case II: Convolution with QAM}\label{sec:QAMSpec}

VecComp offers an innovative approach to convolution operations, a cornerstone in convolutional neural networks (CNNs). Traditionally, CNNs have been implemented using over-the-air analog computation, as explored in~\cite{sanchez2022airnn}, where the convolution operation is achieved through reconfigurable intelligent surfaces. Contrary to this analog-based method, VecComp facilitates convolution operations via digital computing. This approach negates the need for complex reconfiguration and transition to analog systems, presenting a more straightforward, digital-centric solution for executing convolution operations in CNNs. More precisely, for a given vector $\bm{a}\in \mathbb{F}_{Q}^{L+K-1}$, the convolution operator of $\bm{a}$ and nodes' data $\bm{s}$ of size $K \times 1$ can be written as 
\begin{align}
\bm{f}(\bm{s}) = \mathcal{H}(\bm{a})\bm{s}, 
\end{align}
where $\mathcal{H}(\bm{a})$ with size of ${L \times K}$ indicates the Hankel transform of vector $\bm{a}$ with matrix pencil parameter $K$~\cite{Razavikia2019Hankel}. Then, the element-wise formula is given by
\begin{align}
f_{ \ell}(s_1,\ldots,s_K) =  \sum\nolimits_{k=1}^K a_{\ell-1+k}s_k,  \quad \ell \in [L].
\end{align}
To compute the given function using QAM, we can solve the optimization problem in~\eqref{eq:feasibility-convex-lift} to obtain the optimal modulation vector. However, one can use a closed-form formula proposed in~\cite[Section III]{razavikia2023SumCode} as a solution to~\eqref{eq:feasibility-convex-lift}. For $\mathbb{F}$ equals to $\mathbb{Z}$, node $k$ can encode $a_{\ell-1+k}s_k$ into in-phase and quadrature components of digital modulation as follows:
\begin{align}
  \nonumber
    \mathscr{E}_{k,\ell}(s) & :=  a_{\ell-1+k}s - {Q} \cdot \left\lfloor\frac{{a_{\ell-1+k}s}}{Q}\right\rfloor + \frac{1-{Q}}{2}\\
    & +  j\bigg(\Big\lfloor \frac{{a_{\ell-1+k}s}}{Q} \Big\rfloor + \frac{1-Q}{2}\bigg),
\end{align}
in which $\mathscr{E}_{k}$ maps the input value onto the constellation points of a \ac{QAM} of order $Q^2$. Having encoders $\mathscr{E}_{k}$'s, the decoder accordingly becomes:
\begin{align*}
    \mathscr{T}_{\ell}(y)=  \mathcal{W} (\mathfrak{Re}{(y)}) + {Q} \cdot \Big(\mathcal{W}(\mathfrak{Im}{(y)}) + \tfrac{{Q} -1}{2}\Big) + \frac{{Q}-1}{2},
\end{align*}
where $\mathcal{W}(\cdot)$ is the round up to half function, i.e., $ \mathcal{W}(z) =  \lceil z + 0.5\rceil - 0.5$. VecComp's QAM-based modulation encoders can accelerate computer applications, particularly in scenarios where vector-based computations are needed for convolutional operations.

\begin{remark}
    The given encoders in subsections~\ref{sec:PAMspec} and~\ref{sec:QAMSpec} particularly work only for affine and convolution functions with corresponding modulations, respectively. For general function computation, $\mathcal{P}_{\ell}^{\rm inexact}$ may not have a closed-form solution, and to perform the computation, we need to follow the procedure in Algorithm~\ref{alg:Ecnoders}.
\end{remark}

In the following section, we assess the performance of the proposed VecComp method for computation matrix function over the fading \ac{MAC}. 

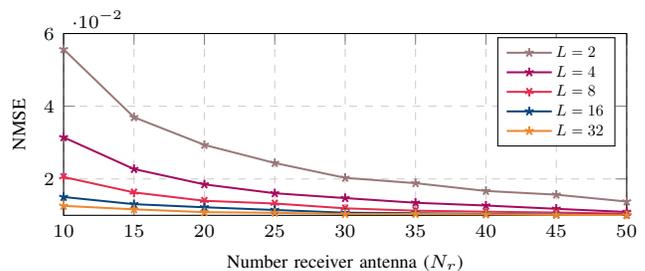
\begin{figure}[!t]
    \centering
 \begin{tikzpicture} 
    \begin{axis}[
        xlabel={Number receiver antenna ($N_r$)},
        ylabel={NMSE},
        label style={font=\scriptsize},
        legend cell align={left},
        tick label style={font=\scriptsize} , 
        width=0.5\textwidth,
        height=4cm,
        xmin=10, xmax=50,
        ymin=1e-2, ymax=6e-2,
       legend style={nodes={scale=0.6, transform shape}, at={(0.98,0.98)}}, 
        ymajorgrids=true,
        xmajorgrids=true,
        grid style=dashed,
        grid=both,
        grid style={line width=.1pt, draw=gray!15},
        major grid style={line width=.2pt,draw=gray!40},
    ]
   \addplot[
        color=bazaar,
        mark=star,
        line width=0.75pt,
        mark size=1.75pt,
        ]
    table[x=Nr,y=L2]
    {Data/MSEErrorNr.dat};
    \addplot[
        color=jazzberryjam,
        mark=star,
        line width=0.75pt,
        mark size=1.75pt,
        ]
    table[x=Nr,y=L4]
    {Data/MSEErrorNr.dat};
    \addplot[
        color=amaranth,
        mark=star,
        mark options = {rotate = 180},
        line width=0.75pt,
        mark size=1.75pt,
        ]
    table[x=Nr,y=L8]
    {Data/MSEErrorNr.dat};
      \addplot[
        color=darkcerulean,
        mark=star,
        line width=0.75pt,
        mark size=1.75pt,
        ]
    table[x=Nr,y=L16]
    {Data/MSEErrorNr.dat};
      \addplot[
        color=cadmiumorange,
        mark=star,
        line width=0.75pt,
        mark size=1.75pt,
        ]
    table[x=Nr,y=L32]
    {Data/MSEErrorNr.dat};
    \legend{$L=2$,$L=4$,$L=8$,$L=16$, $L=32$};
    \end{axis}
\end{tikzpicture}

\caption{NMSE performance of VecComp as a function of the number of receiver antennas, $N_r$, for varying computation dimensions $L$. The setup assumes $N_t = L $ and $\text{SNR} = 20, \text{dB}$, with simulation results averaged over $10^3$ Monte Carlo trials. The figure illustrates the improvement in computation robustness and fading resilience as $N_r$ increases from $10$ to $50$, highlighting the effect of antenna scaling on VecComp’s ability to accurately compute the sum function in a fading environment.}

\label{fig:NumberAnteenaNr}
\end{figure}

\begin{figure}[!t]
    \centering
    \begin{tikzpicture} 
        \begin{axis}[
            xlabel={Number of nodes ($K$)},
            ylabel={NMSE},
            label style={font=\scriptsize},
            legend cell align={left},
            tick label style={font=\scriptsize},
            width=0.5\textwidth,
            height=4cm,
            xmin=4, xmax=32,
            ymin=0.56, ymax=0.6,
            legend style={nodes={scale=0.6, transform shape}, at={(0.98,0.98)}},
            ymajorgrids=true,
            xmajorgrids=true,
               minor tick num=5,
            grid style=dashed,
            grid=both,
            grid style={line width=.1pt, draw=gray!15},
            major grid style={line width=.2pt,draw=gray!40},
        ]
        \addplot[
            color=bazaar,
            mark=star,
            line width=0.75pt,
            mark size=1.75pt,
        ]
        table[x=K,y=Nr128]
        {Data/MSEErrorNodes.dat};
        \addplot[
            color=jazzberryjam,
            mark=star,
            line width=0.75pt,
            mark size=1.75pt,
        ]
        table[x=K,y=Nr512]
        {Data/MSEErrorNodes.dat};
        \addplot[
            color=amaranth,
            mark=star,
            mark options = {rotate = 180},
            line width=0.75pt,
            mark size=1.75pt,
        ]
        table[x=K,y=Nr2048]
        {Data/MSEErrorNodes.dat};
        \legend{$N_r=128$,$N_r=512$,$N_r=2048$};
        \end{axis}
    \end{tikzpicture}
    \caption{ NMSE performance of VecComp as a function of the number of transmitting nodes, $K$, for different receiver antenna configurations. The simulation is conducted with $N_t=4$, $L=4$, and $\text{SNR}=5$~dB, with results averaged over $10^4$ Monte Carlo trials.}
    \label{fig:NumberNodes}
\end{figure}
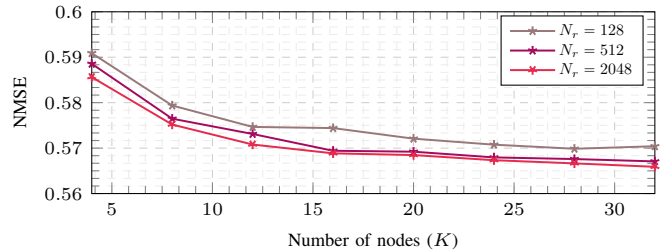

\section{Numerical Experiments}\label{sec:num}

{ \change
We divide this section into three parts to evaluate VecComp's performance for fading compensation and vector computation tasks for different numbers of nodes, antennas, and noise levels. In the first part,  we analyze the performance of the proposed beamforming technique to compensate for the fading effect. Next, we analyze VecComp's performance for computing various functions with multiple orders of modulations and several nodes over the noisy MAC.  Finally, we evaluate the computation performance over different noise levels. In particular,  we compare the performance of VecComp on the following three settings: $1)$ an analog MIMO approach \cite{zhu2018mimo} in which the sum function is computed over the air, while general functions are obtained via nomographic approximation. $2)$ a conventional digital MIMO transmission scheme based on}{ \change orthogonal frequency-division multiple access, where nodes occupy distinct frequency subchannels and multiple function outputs are supported through MIMO processing; and $3)$ the digital SumComp scheme~\cite{razavikia2023SumCode}, which is extended in this work to a MIMO setup for the simulations and follows a structure similar to the analog MIMO approach but employs digital QAM modulation for sum computation and nomographic approximation for general functions.
To maintain focus on VecComp's primary objective, which is achieving reliable vector computation in fading environments, we mainly concentrate on function computation and channel compensation within the VecComp framework. }

\subsection{Number of Antennas}

In this subsection, we explore how the number of antennas impacts VecComp’s ability to compensate for fading effects and accurately compute the sum function. The nodes compute the sum function over fading \ac{MAC}, where channel coefficients are generated randomly according to the Gaussian distribution, i.e., $\bm{H}_k\sim \mathcal{CN}(\bm{0}, N_r\sigma_{h}^2\bm{I}_{N_t})$ with $\sigma_h^2=1$.  For beamforming vectors, we generated them by $\bm{V}_k\sim \mathcal{CN}(\bm{0}, \bm{I}_{L})$ to ensure that they satisfy the isotropic properties. We also consider scenarios where the number of nodes $K=100$, the number of transmitter antennas $N_t=L$, and the number of receiver antennas varies, studying configurations ranging from $N_r=10 $ to $N_r=50 $ to determine the effect of antenna scaling on normalized MSE (NMSE) performance. The simulation results averaged over $10^3$ Monte Carlo trials with signal-to-noise ratios (SNRs) equal to $20$ dB. Figure~\ref{fig:NumberAnteenaNr} illustrates how the increase in receiver antenna $N_r$, from $10$ to $50$, improves computation robustness and fading resilience across various function computation $L$. Moreover, the increased number of functions for computation  $L$ leads to lower NMSE. In particular, by increasing the number of antennas from $N_r =10$ to $N_r=50$, we observe around $\% 75$ reduction of NMSE.

\begin{figure}[!t]
    \centering
    \begin{tikzpicture}
        \begin{axis}[
            xlabel={SNR (dB)},
            ylabel={NMSE},
            ymode=log,
            width=0.5\textwidth,
            height=4cm,
            xmin=-5, xmax=25,
            legend cell align={left},
            legend style={nodes={scale=0.6, transform shape}, at={(0.98,0.98)}},
            tick label style={font=\scriptsize},
            label style={font=\scriptsize},
            grid=both,
               minor tick num=5,
            grid style={line width=.1pt, draw=gray!15},
            major grid style={line width=.2pt,draw=gray!40},
        ]
            \addplot[
                color=jazzberryjam,
                mark=star,
                line width=0.75pt,
                mark size=1.75pt,
            ]
            table[x=SNR,y=VecComp] {Data/NMSE_Agg_Mod.dat};
            
            \addplot[
                color=amaranth,
                mark=star,
                mark options={rotate=180},
                line width=0.75pt,
                mark size=1.75pt,
            ]
            table[x=SNR,y=Analog] {Data/NMSE_Agg_Mod.dat};
            
            \addplot[
                color=darkcerulean,
                mark=star,
                mark options={rotate=180},
                line width=0.75pt,
                mark size=1.75pt,
            ]
            table[x=SNR,y=OFDMA] {Data/NMSE_Agg_Mod.dat};
        \addplot[
                color=cadmiumgreen,
                mark=star,
                mark options={rotate=180},
                line width=0.75pt,
                mark size=1.75pt,
            ]
            table[x=SNR,y=SumComp] {Data/NMSE_Agg_Mod.dat};
            
            \legend{VecComp, MIMO OAC~\cite{zhu2018mimo}, Wideband MIMO, MIMO SumComp~\cite{razavikia2023SumCode}};
        \end{axis}
    \end{tikzpicture}
    \caption{ \change Performance comparison between VecComp, MIMO OAC~\cite{zhu2018mimo}, MIMO SumComp~\cite{razavikia2023SumCode}, and wide-band MIMO in terms of NMSE error averaged over $N_s =100$, where input values are given by $x_k=\{0,1,\ldots,7\}$  and the desired functions are $f_1 = \prod_{k}x_k$, $f_2 = \sum_{k}x_k/K$, $f_3 = \max_{k}x_k$, and $f_4 = \sum_{k}x_k^2$.}
    \label{fig:NMSEPerformance}
\end{figure}
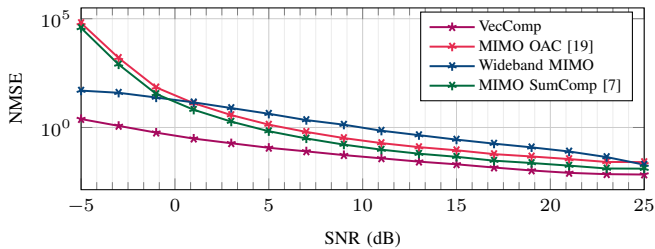

\subsection{Number of Nodes}
In this subsection, we assess the effect of varying the number of transmitting nodes on the performance of  VecComp. The experiment considers a variable number of nodes, $K$, ranging from $4$ to $32$ in increments of $4$, while fixing the number of transmitter antennas at $N_t=4$ and the beamforming dimension at $L=4$. Consistent with the previous analysis, channel matrices are generated as $\bm{H}_k\sim \mathcal{CN}(\bm{0}, N_r\sigma_h^2\bm{I}_{N_t})$ with $\sigma_h^2=1$, and beamforming matrices are drawn from $\bm{V}_k\sim \mathcal{CN}(\bm{0}, \bm{I}_{L})$, normalized to satisfy the power constraints. The combined received signal, resulting from the superposition of contributions from all nodes and subject to additive noise, is processed to yield the NMSE as the performance metric.
\begin{figure*}[t]
\definecolor{eggplant}{rgb}{0.38, 0.25, 0.32}

\centering
\pgfplotsset{every axis/.append style={
    xlabel style={font=\tiny},
    ylabel style={font=\tiny},
    tick label style={font=\tiny},
    title style={font=\tiny},
    ymode=log,
    width=0.39\textwidth,
    height=4cm,
    ymajorgrids=true,
    xmajorgrids=true,
    xmin =2,
    xmax = 16,
    ymin = 0.05,
    grid=both,
    grid style={line width=.1pt, draw=gray!15},
    major grid style={line width=.2pt, draw=gray!40},
     minor tick num=5,
}}
\begin{tikzpicture}
  \pgfmathsetmacro{\Kfixed}{32}   
  \pgfmathsetmacro{\qfixed}{64}   
  \pgfmathsetmacro{\Lfixed}{4}    

  \begin{groupplot}[
    group style={group size=3 by 1, horizontal sep=8pt},
  ]

  \nextgroupplot[
   xlabel={\footnotesize $q, (K,L)=(32,4)$},
    ylabel={\footnotesize Computational Complexity},
  ]
  \addplot+[very thick,bluegray, solid, mark=triangle, domain=2:16, samples=20] {(\Lfixed)};
  \addplot+[very thick, densely dashed, jazzberryjam,mark=none, domain=2:16, samples=20]
    { (max(x,\Kfixed))^(4*(\Lfixed-1)) * x^(0.5*(\Lfixed-1)) };
  \addplot+[thick, solid,eggplant,  mark=star, domain=2:16, samples=20] {(\Lfixed)};
  \addplot+[thick, darkcerulean, densely dashed, domain=2:16, samples=20]
    { (\Kfixed*x)^( (\Lfixed-1) ) };
\addplot+[very thick,cadmiumgreen, mark=o, domain=2:16, samples=20] {2};
  \nextgroupplot[
   xlabel={\footnotesize  $K, (q,L)=(64,4)$},
   yticklabels={},
ylabel={},
  ]
  \addplot+[very thick, solid,bluegray, mark=triangle, domain=2:16, samples=20] {(\Lfixed)};
  \addplot+[very thick, densely dashed,jazzberryjam, mark=none, domain=2:16, samples=20]
    { (max(x,\qfixed))^(4*(\Lfixed-1)) * \qfixed^(0.5*(\Lfixed-1)) };
  \addplot+[thick,  solid, eggplant,  mark=star, domain=2:16, samples=20] {(\Lfixed)};
  \addplot+[thick, darkcerulean, densely dashed , domain=2:16, samples=20]
    { (x*\qfixed)^( (\Lfixed-1) ) };
\addplot+[very thick,cadmiumgreen, mark=o, domain=1:16, samples=16] {2};
  \nextgroupplot[
    xlabel={\footnotesize  $L, (q,K)=(64,32)$},
    yticklabels={},
ylabel={},
    legend style={font=\tiny, at={(0.55,0.26)}, anchor=south east}
  ]
  \addplot+[very thick, solid,bluegray, mark=triangle, domain=1:16, samples=16] {x};
  \addplot+[very thick, densely dashed,jazzberryjam, mark=none, domain=1:16, samples=16]
    { (max(\qfixed,\Kfixed))^(4*(x-1)) * \qfixed^(0.5*(x-1)) };
  \addplot+[thick,  solid,eggplant,  mark=star, domain=1:16, samples=16] {x};
  \addplot+[thick, darkcerulean, densely dashed, domain=1:16, samples=16]
    { (\Kfixed*\qfixed)^(x-1) };
\addplot+[very thick,cadmiumgreen, mark=o, domain=1:16, samples=16] {x/4};

  \legend{
    SDP: VecComp, SDP: Na\"ive-ChannelComp,
    Dec: VecComp, Dec: Na\"ive-ChannelComp, Dec:MIMO SumComp
  }
  \end{groupplot}
\end{tikzpicture}
\caption{ \change
Computational complexity comparison of VecComp, MIMO SumComp,  and a naïve $L$-dimensional extension of ChannelComp. The figure depicts the ratio of the complexities, both for solving the SDP in the modulation design and for the decoding procedure, relative to ChannelComp. Subplots from left to right vary with  modulation order $q$,  number of nodes $K$, and vector length $L$, respectively. The logarithmic scale highlights the linear growth of VecComp in $L$ versus the exponential growth of the naïve scheme.}

\label{fig:ratio_sdp_dec}
\end{figure*}
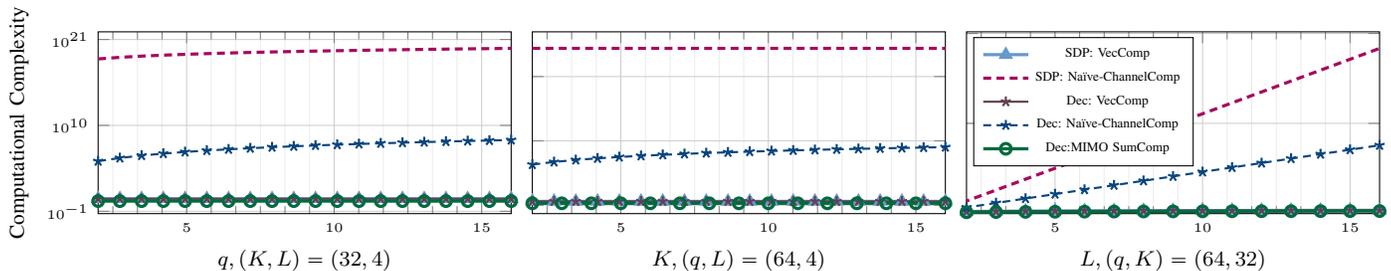

Figure~\ref{fig:NumberNodes} illustrates that the NMSE remains approximately constant as $K$ increases, which is consistent with the prior experiment where increasing the number of antennas reduced the overall error. Although the MSE increases with the number of nodes—owing to the higher aggregate power in $\|\sum_k \bm{x}_k\|$—this effect is counterbalanced by the corresponding increase in the norm of the sum of transmitted symbols as elucidated by Remark~\ref{rem:nodes}. Consequently, the NMSE, defined as the MSE normalized by $\|\sum_k \bm{x}_k\|^2$, remains invariant to $K$. This result underscores the robustness of the VecComp scheme in accommodating a massive number of nodes.

\subsection{Complex Modulations}

{ \change
 We assess the impact of VecComp's designed modulation on the aggregated function computation performance over the noisy MAC compared to the following methods: 1) analog MIMO scheme~\cite{zhu2018mimo}, where nodes use analog modulation for transmission, 2) wideband MIMO scheme, where each node transmits over a distinct communication channel and uses multiple antennas to compute multiple functions at the CP, and 3) digital MIMO scheme~\cite{razavikia2023SumCode}, termed SumComp, where nodes use QAM style modulation for transmission and computing sum function over the MAC.  Note that we are only comparing the computation aspect of the mentioned scheme here, as there is no fading. Four distinct functions, $L=4$, are considered: the product, mean, maximum, and sum-of-squares, each realized via a dedicated modulation vector obtained from \eqref{eq:feasibility-convex-lift}.  The NMSE is computed over a range of SNR values from $-5$ dB to $25$~dB.

Figure~\ref{fig:NMSEPerformance} illustrates the NMSE performance for the three methods as a function of SNR. The results demonstrate that, as the SNR increases, the aggregated NMSE decays monotonically. 
Moreover, VecComp shows superior performance thanks to the designed modulation for each desired function, which improved around $10$ dB in NMSE in low SNR areas. Furthermore, the analog MIMO scheme cannot accurately compute the maximum function, even in low SNR scenarios, due to its approximation techniques for computing the maximum and product functions.

}

{ \change
\subsection{Computational Complexity of the Design}

To complement the performance results, we provide the computational complexity of {VecComp}, MIMO SumComp~\cite{razavikia2023SumCode},  and a naïve $L$-dimensional extension based on Remark~\ref{rem:complexity}, both relative to ChannelComp. 
Figure~\ref{fig:ratio_sdp_dec} depicts  the ratios of the SDP solving cost and decoding cost with respect to {ChannelComp} when $L=1$, while varying the order of modulation $q$, the number of nodes $K$, and the vector length $L$. 
The results show that the complexity of {VecComp} grows only linearly in $L$, leading to a constant ratio (equal to $L$) when compared }{\change with {ChannelComp}. In contrast, MIMO SumComp does not incur any optimization overhead, as it admits a closed-form solution for modulation mapping. The naïve extension of ChannelComp, however, exhibits exponential growth in $L$, with ratios that increase rapidly on the logarithmic scale. 
These results highlights  that {VecComp} achieves substantial computational savings: it preserves the same order of complexity as {ChannelComp} in $q$ and $K$, while scaling with a linear factor $L$.  
}

\subsection{Computation Performance for Affine Transform}

\begin{figure}[t]
    \centering

 \begin{tikzpicture} 
    \begin{axis}[
        xlabel={SNR (dB)},
        ylabel={NMSE},
        label style={font=\scriptsize},
        legend cell align={left},
        tick label style={font=\scriptsize} , 
        width=0.5\textwidth,
        height=4cm,
        xmin=10, xmax=30,
        ymode = log,
       legend style={nodes={scale=0.6, transform shape}, at={(0.98,0.98)}}, 
        ymajorgrids=true,
        xmajorgrids=true,
        grid style=dashed,
        grid=both,
        grid style={line width=.1pt, draw=gray!15},
        major grid style={line width=.2pt,draw=gray!40},
    ]
    \addplot[
        color=jazzberryjam,
        mark=star,
        line width=0.75pt,
        mark size=1.75pt,
        ]
    table[x=SNR,y=Q8]
    {Data/NMSEvsKQAM.dat};
    \addplot[
        color=amaranth,
        mark=star,
        mark options = {rotate = 180},
        line width=0.75pt,
        mark size=1.75pt,
        ]
    table[x=SNR,y=Q16]
{Data/NMSEvsKQAM.dat};
        \addplot[
        color=darkcerulean,
        mark=star,
        mark options = {rotate = 180},
        line width=0.75pt,
        mark size=1.75pt,
        ]
    table[x=SNR,y=Q32]
    {Data/NMSEvsKQAM.dat};
    \legend{PAM $64$, PAM $256$, PAM $1024$};
    \end{axis}
\end{tikzpicture}

\caption{Performance of adapting PAM modulations over the noisy MAC.  The function is the affine transform in \eqref{eq:affine_function} with $K=10$ and $L = 10$, where coefficients  $a_{\ell,k}$  are generated randomly from $\{0,1,\ldots, Q-1\}$. The figure shows the performance for different SNRs and the moderation order when the ${\rm SNR} = 10$ dB over $5 \times 10^3$ Monte Carlos trials.}
\label{fig:QAMExp}
\end{figure}
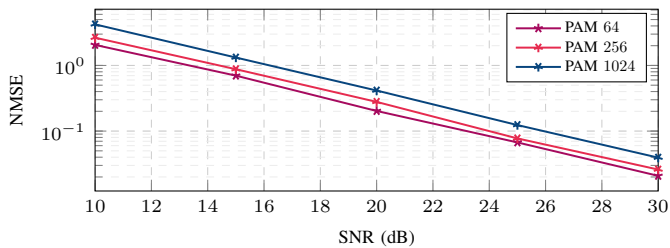

We study the performance of the computation over the noisy \ac{MAC}, for computing an affine function in \eqref{eq:affine_function}, where $a_{k,\ell}$ are generated randomly from the set  $\{0,1,\ldots, Q-1\}$  for all $(k,\ell)\in [K]\times [L]$ with $K=50$ nodes. Furthermore, the input data $\bm{s}$ is selected uniformly at random from the integers number between $0$ and $Q-1$, i.e.,  $s_{k,\ell} =\{0,1,\ldots, Q-1\}$ for all $(k,\ell)\in [K]\times [L]$. We use the PAM as described in \ref{sec:PAMspec} for the modulation. The results are averaged over $5\times 10^3$ Monte Carlo trials. In Figure~\ref{fig:QAMExp}, we depict the performance of the VecComp over various numbers of nodes $K$ in terms of NMSE for PAM of orders $Q= \{4,8,16,32\}$ with $L=5$.  By increasing the order of modulation $Q$, the MSE increases due to reducing the distance of the constellation points. 

\subsection{Computation Performance for Convolution}

For the last experiment, we evaluate the computation performance of VecComp for different SNRs and modulation orders. For the function computation, we consider the special scenario described in Section~\ref{sec:QAMSpec}, where the desired function is the convolution, and the nodes use QAM for sending their values. This analysis examines the accuracy and robustness of the VecComp approach under various ranges of input values levels $s_{k,\ell}$ in $\{0,1,\ldots, Q\}$, where \( Q \in \{4, 8, 16\} \) corresponds to QAM of order $16,64$ and $256$, respectively. The SNR values tested range from $10$ to $30$ dB, allowing us to assess how noise and modulation complexity affect NMSE in the computation.

\begin{figure}
    \centering

 \begin{tikzpicture} 
    \begin{axis}[
        xlabel={SNR (dB)},
        ylabel={NMSE},
        label style={font=\scriptsize},
        legend cell align={left},
        tick label style={font=\scriptsize} , 
        width=0.5\textwidth,
        height=4cm,
        xmin=5, xmax=30,
        ymode = log,
       legend style={nodes={scale=0.6, transform shape}, at={(0.98,0.98)}}, 
        ymajorgrids=true,
        xmajorgrids=true,
        grid style=dashed,
        grid=both,
        grid style={line width=.1pt, draw=gray!15},
        major grid style={line width=.2pt,draw=gray!40},
    ]
   \addplot[
        color=bazaar,
        mark=star,
        line width=0.75pt,
        mark size=1.75pt,
        ]
    table[x=SNR,y=Q16]
    {Data/MSEQAMr.dat};
    \addplot[
        color=jazzberryjam,
        mark=star,
        line width=0.75pt,
        mark size=1.75pt,
        ]
    table[x=SNR,y=Q64]
    {Data/MSEQAMr.dat};
    \addplot[
        color=amaranth,
        mark=star,
        mark options = {rotate = 180},
        line width=0.75pt,
        mark size=1.75pt,
        ]
    table[x=SNR,y=Q256]
    {Data/MSEQAMr.dat};
    \legend{ {QAM $16$}, {QAM $64$}, {QAM $256$}};
    \end{axis}
\end{tikzpicture}

\caption{NMSE performance of VecComp for different QAM modulation orders \( Q^2 \in \{16, 64, 256\} \) and SNR levels ranging from $5$ dB to $30$ dB. The results, averaged over $10^4$ Monte Carlo trials, illustrate how increasing SNR improves computation accuracy across all modulation levels, while higher modulation orders (e.g., 256-QAM) show increased MSE due to their higher sensitivity to noise.}

\label{fig:QAMSpecial2}
\end{figure}
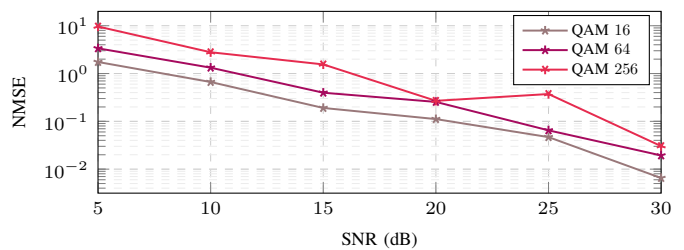

In each trial, we simulate the convolution operation over the noisy MAC, applying QAM modulation and demodulation to the convolution results over the noisy MAC. Also,  the experiments are averaged over $10^4$ Monte Carlo trials. Results are depicted in Figure~\ref{fig:QAMSpecial2}, showing the impact of SNR on MSE across different QAM modulation orders. The higher modulation orders (e.g., $256$-QAM) yield higher NMSEs due to their increased susceptibility to noise. However, as SNR increases from  $10$ to $30$ dB, MSE decreases across all modulation levels, demonstrating that higher SNR conditions improve VecComp’s accuracy and computation robustness as expected. These results confirm the trade-offs between noise resilience and modulation complexity in the digital computation framework of VecComp.

\section{Conclusion}\label{sec:conclusion}

This paper introduced VecComp, an extension of the ChannelComp methodology that leverages MIMO technology to enable robust vector-based computations over fading channels. Our analysis established a non-asymptotic upper bound on the MSE of VecComp, supporting its theoretical efficacy in mitigating fading effects. Numerical experiments confirmed that VecComp effectively balances computation accuracy and resilience across various modulation orders and SNR levels. Specifically, the scaling of receiver antennas and modulation order demonstrated the adaptability of VecComp in achieving computational reliability in data-centric applications requiring vector-based computations. Furthermore, the versatility of VecComp renders it applicable to a broad range of practical scenarios.  Potential applications include real-time distributed machine learning, large-scale sensor networks, IoT deployments, and task-oriented OAC,  where efficient  OAC is paramount.

These findings indicate that VecComp is a promising solution for enhancing computational and communication reliability in digital over-the-air vector function computations, marking a step toward scalable, reliable, and efficient distributed data processing.


\appendix

\subsection{Proof of Theorem~\ref{th:NError}}\label{sec:ProofTheorem}

Consider the target signal $\bm{r}: = \sum_{k=1}^K\bm{x}_k $, we can define the error as follows:
\begin{align}
       \Big\|\bm{r} - \bm{y}\Big\|_2 & = \Big\| \bm{r} - \bm{y}_{\rm sig} - \bm{y}_{\rm inter} - \bm{y}_{\rm noise}\Big\|_2, \nonumber \\
      \label{eq:MIMO-error} & \leq  e_{\rm sig} + e_{\rm inter} + e_{\rm noise},
\end{align}
where 
\begin{subequations}
\label{eq:Errros}
\begin{align}
 e_{\rm sig} & : =  \Big\| \bm{r} - \frac{1}{N_r} \sum\nolimits_{k=1}^K\bm{G}_k^{\mathsf{H}}\bm{G}_k\bm{x}_k\Big\|_2, \\
e_{\rm inter}  & := \Big\| \frac{1}{N_r} \sum\nolimits_{k,k',k\neq k' }^K\bm{G}_{k'}^{\mathsf{H}}\bm{G}_k\bm{x}_k \Big\|_2,  \\
  e_{\rm noise}& :=  \Big\| \frac{1}{N_r} \sum\nolimits_{k=1}^K\bm{G}_k^{\mathsf{H}}\bm{z}\Big\|_2,  
\end{align}
\end{subequations}
with $\bm{G}_k:= \bm{H}_k\bm{V}_k/N_t \in \mathbb{C}^{N_r \times L}$ is a random matrix with i.i.d. entries distributed as $\mathcal{CN}({0},1)$.  
We analyze each term separately in the sequel and find how fast the tails of the error terms, $e_{\rm sig}, e_{\rm inter},$ and $e_{\rm noise}$ approach zero. To this end, we use the following concentration inequality for the matrix elements in error terms.


\begin{lem}\label{lem:Bern}(Matrix Bernstein)\cite[Theorem 1.6.2]{tropp2015introduction}, Let $\bm{S}_1, \ldots, \bm{S}_n$ be independent, centered random matrices with common dimension $d_1 \times d_2$, and assume that each matrix is uniformly bounded as
\begin{align}
    \mathbb{E}[\bm{S}_k] = \bm{0}, \quad \|\bm{S}_k\| \leq L \quad, {\rm for~each~}k = 1, \ldots, n.  
\end{align}
Let us introduce the sum $\bm{Z} = \sum_{k=1}^n\bm{S}_k,$ and let $v(\bm{Z})$ denote the matrix variance statistic of the sum, defined as
\begin{align*}
     v(\bm{S})&  = \max\{\|\mathbb{E}[\bm{Z}\bm{Z}^{\mathsf{H}}] \|, \|\mathbb{E}[\bm{Z}^{\mathsf{H}}\bm{Z}] \| \} \\
     & = \max\Big\{\Big\|\sum\nolimits_{k=1}^n\mathbb{E}[\bm{S}_k\bm{S}_k^{\mathsf{H}}] \Big\|, \Big\|\sum\nolimits_{k=1}^n\mathbb{E}[\bm{S}_k^{\mathsf{H}}\bm{S}_k] \Big\| \Big\}.
\end{align*}
Then, 
\begin{align*}
    \mathbb{P}\big[\|\bm{Z}\| \geq t\big] \leq (d_1 + d_2) \exp{\Big(\frac{-t^2/2}{v(\bm{Z}) + Lt/3}\Big)}  \quad {\rm for~all}~ t \geq 0 
\end{align*}
\end{lem}
\begin{cor}\label{cor:Bern}(Expectation Upper bound)\cite[Section 1.6]{tropp2015introduction}
Furthermore,  the expectation of the spectral norm of the matrix $\bm{Z}\in \mathbb{R}^{d_1\times d_2}$ can be bound as
\begin{align}
     \mathbb{E}[\|\bm{Z}\|] \leq \sqrt{2v(\bm{Z})\log(d_1+d_2)} + \frac{1}{3}L\log{(d_1+d_2)}.
\end{align}
\end{cor}

In the following, we use the concentration inequalities in Lemma~\ref{lem:Bern} to obtain a probabilistic upper bound on each error term in~\eqref{eq:Errros}. Consequently, we can show that the upper bound for each error term asymptotically reaches zero for a large enough number of antennas, $N_r$.

\begin{lem}\label{lem:Delta1}
    Let $\epsilon_1>0$ and $\delta_1>0$ be positive scalars. Then,  the absolute value of the signal error  $|e_{\rm sig}|$ is upper bounded by the scalar $\epsilon_1$, with probability at least $1-\delta_1$ as follows:
\begin{align}
 |e_{\rm sig}| \leq \epsilon_1,~\text{if}~ N_r
\ge
\frac{2\gamma L}{\epsilon_1^{2}}
\ln\!\Big(\frac{L+1}{\delta_1}\Big),,
\end{align}
    where $\gamma = \sum_{k=1}^{K}\|\bm{x}_{k}\|_2^2$. 
\end{lem}
\begin{proof}
    See Appendix~\ref{sec:lemDelta1}.
\end{proof}

\begin{lem}\label{lem:Delta2}
    Let $\epsilon_2>0$ and $\delta_2>0$ be positive scalars. Then,  the absolute value of the interference error,  $|e_{\rm inter}|$,  is upper bounded by the scalar $\epsilon_2$, with probability at least $1-\delta_2$ as follows: 
\begin{align}
   |e_{\rm inter}| \leq \epsilon_2~\text{if}~ N_r \geq  \frac{2\gamma LK^2}{\epsilon_2^2}\ln{\Big(\frac{K(L+1)}{\delta_2}\Big)}.
\end{align}
\end{lem}
\begin{proof}
    See Appendix~\ref{sec:lemDelta2}.
\end{proof}

\begin{lem}\label{lem:Delta3}
    Let $\epsilon_3>0$ and $\delta_3>0$ be positive scalars. Then,  the abaloute value of the noise error,   $|e_{\rm noise}^n|$, is upper bounded by the scalar $\epsilon_3$  with probability at least $1-\delta_3$ as follows
\begin{align}
   |e_{\rm noise}^n| \leq \epsilon_3~\text{if}~ N_r \geq  \frac{8L K \sigma_z^2 }{\epsilon_3^2}\ln{\Big(\frac{L+1}{\delta_3}\Big)}.
\end{align}
\end{lem}
\begin{proof}
    See Appendix~\ref{sec:lemDelta3}.
\end{proof}
The proofs of Lemmas~\ref{lem:Delta1},~\ref{lem:Delta2}, and~\ref{lem:Delta3} are straightforward by applying Lemma~\ref{lem:Bern} on the corresponding  error term in \eqref{eq:Errros}. Hence, the proofs are omitted for brevity.

By invoking Lemmas~\ref{lem:Delta1},~\ref{lem:Delta2}, and~\ref{lem:Delta3}, we obtain an upper bound on the error terms in~\eqref{eq:MIMO-error} below: 
\begin{align}
\label{eq:error_union}
    \|\bm{e}\|_2 \leq e_{\rm sig} + e_{\rm inter} + e_{\rm noise} \leq \epsilon_1 +  \epsilon_2 +  \epsilon_3 \leq \epsilon.
\end{align}
Due to the union bound, the events of Lemmas~\ref{lem:Delta1},~\ref{lem:Delta2}, and~\ref{lem:Delta3} hold simultaneously with probability no less than $1-(\delta_1 + \delta_2 +\delta_3)$ or $1-\delta$, where $\delta \geq \delta_1 + \delta_2 +\delta_3$~\cite{Razavikia2019Hankel}. Thus, this proves~\eqref{eq:epsilonupp}. 

Then,  all the Lemmas can be satisfied if the number of antennas at the \ac{CP} is large enough, i.e., 
\begin{align}
 \nonumber
    N_r \geq  2\max\Bigg\{&  \frac{\gamma L}{\epsilon_1^2}\ln{\Big(\frac{L+1}{\delta_1}\Big)}, \frac{\gamma LK^2}{\epsilon_2^2}\ln{\Big(\frac{K(L+1)}{\delta_2}\Big)}, \\
    & \frac{4L{K}\sigma_z^2}{\epsilon_3^2}\ln{\Big(\frac{L+1}{\delta_3}\Big)} \Bigg\}.
\end{align}
 By setting $\delta_1 = \delta/2K, \delta_2 = \delta/2$, and $\delta_3 = \delta/2K$, we obtain the following lower bound: 
\begin{align}
    \label{eq:Nrfirst}
       \hspace{-5pt} N_r \geq 2 \max\Bigg\{ \frac{\gamma L}{\epsilon_1^2}, \frac{\gamma LK^2}{\epsilon_2^2}, \frac{4L{K}\sigma_z^2}{\epsilon_3^2}\Bigg\}\ln{\Big(\frac{2K(L+1)}{\delta}\Big)},
\end{align}
where $\delta >0$, and for any $K\geq 2$,  we have
\begin{align}
    \delta_1 + \delta_2 + \delta_3 = \Big(\frac{1}{2}+ \frac{1}{2K} + \frac{1}{2K}
     \Big) \delta \leq \delta.  
\end{align}
For further simplifications, we reformulate $\epsilon_1, \epsilon_2$, and $\epsilon_3$ in terms of a positive value $\epsilon'>0$ as follows: 
\begin{align}
\label{eq:Cconstant}
    \epsilon_1  =  \frac{\epsilon'}{K{\sigma_z}},~~
    \epsilon_2  = \frac{1}{{\sigma_z}}\epsilon' ,~~
    \epsilon_3  = \frac{2\epsilon'}{\gamma\sqrt{K}}.
\end{align}
Hence, it yields the following: {\change
\begin{align}
\nonumber
    \epsilon_1 + \epsilon_2+ \epsilon_3 
    & \nonumber
    \leq  {\epsilon'}\Big( \frac{1}{K{\sigma_z}} +  \frac{1}{{\sigma_z}} + \frac{2}{\gamma\sqrt{K}}
    \Big), \\\nonumber
    & \leq  \epsilon'\Big( \frac{2}{{\sigma_z}} + \frac{2}{{\gamma}})\leq \frac{\epsilon'}{{\min\{{\sigma_z}/4, \gamma /4\}}}= \frac{\epsilon'}{\sqrt{c_{\sigma, \gamma}}}, 
\end{align}
where $c_{\sigma, \gamma}:= \min\{{\sigma_z^2}, \gamma^2 \}/16$. Substituting~\eqref{eq:Cconstant} into~\eqref{eq:Nrfirst}, we obtain 
\begin{align}
 \label{eq:proofnr}
    N_r \geq  \frac{2c_{\sigma, \gamma} LK^2\gamma \sigma_z^2}{\epsilon'^{2}}\ln{\Big(\frac{2K(L+1)}{\delta}\Big)},
\end{align}
 with probability no less than $1-\delta$.  Hence, we conclude the proof.  }


\subsection{Proof of Proposition~\ref{prop:SolP1}}\label{sec:ProofP1prob}
We need to prove that $\hat{\bm{X}}_{\ell} = (\hat{\bm{Y}}_{\ell} +\Tilde{\bm{X}}_{\ell})/\kappa$ satisfies all the constraints in~\eqref{eq:feasibility1}. For the power constraint, we note that
\begin{align}
    \|\hat{\bm{X}}_{\ell} \|_{\rm F}^2 = \frac{1}{\kappa^2} ( \|\hat{\bm{Y}}_{\ell}\|_{\rm F}^2 + \|\tilde{\bm{X}}_{\ell}\|_{\rm F}^2) = 1.
\end{align}
where the first equality is due to the fact that $\langle \tilde{\bm{X}}_{\ell}, \hat{\bm{Y}}_{\ell} \rangle = 0$ for any $\tilde{\bm{X}}_{\ell} \in \mathcal{N}(\mathcal{A}^{\Omega}_{\ell})$ and $\hat{\bm{Y}}_{\ell} \notin \mathcal{N}(\mathcal{A}^{\Omega}_{\ell})$. Next, for the other constraint in~\eqref{eq:feasibility1}, we only need to show that Algorithm~\ref{Alg:non-orthogonal}
gives a non-orthogonal vector to the set of input vectors in the sequel. Let $\{\bm{v}_1,\ldots,\bm{v}_m\}$ be a set of given vectors in $\mathbb{R}^{n}$ and $\bm{x}= {\rm NonOrth}(\bm{v}_1,\ldots,\bm{v}_m)$ be the output from Algorithm~\ref{Alg:non-orthogonal}. For $i\in \{2,\ldots, K\}$, we have  
\begin{align}
    \langle \bm{x}, \bm{v}_i \rangle = \langle \alpha \bm{v}_1+ \bm{y}, \bm{v}_i \rangle = \langle  \bm{y}, \bm{v}_i \rangle \neq 0,
\end{align}
which holds due to the definition of $\bm{y}$ in Algorithm~\ref{Alg:non-orthogonal}. On the other hand, for $i\in \{k+1,\ldots, m\}$, we have $\langle \bm{x}, \bm{v}_i \rangle = \alpha \langle  \bm{v}, \bm{v}_i \rangle  + \langle  \bm{y}, \bm{v}_i \rangle \neq 0$. The last term cannot be zero because then we would have $  \alpha = { -\langle  \bm{y}, \bm{v}_i \rangle}/{\langle  \bm{v}_1, \bm{v}_i \rangle}.$ This contradicts how we selected the parameter $\alpha$ in Algorithm~\ref{Alg:non-orthogonal}. Hence, $\bm{x} \notperp \{\bm{v}_1,\ldots,\bm{v}_m\}$.

\subsection{Proof of Lemma~\ref{lem:Delta1}}\label{sec:lemDelta1}

For each $k\in[K]$,  we define  the random variable $\bm{S}_k\in \mathbb{C}^{L\times 1}$
\begin{equation}
\bm{S}_k := \Big(\bm{I}_L-\frac{1}{N_r}\bm{G}_k^{\mathsf H}\bm{G}_k\Big)\bm{x}_k.
\end{equation}
Since $\mathbb{E}\!\left[\bm{G}_k^{\mathsf H}\bm{G}_k\right]=N_r\bm{I}_L$, it follows immediately that
$\mathbb{E}[\bm{S}_k]=\bm{0}$. Moreover, using the definition of $\bm r$, we have
\begin{equation}
\sum\nolimits_{k=1}^{K}\bm{S}_k
=
\bm r-\frac{1}{N_r}\sum\nolimits_{k=1}^{K}\bm{G}_k^{\mathsf H}\bm{G}_k\bm{x}_k .
\end{equation}

In order to invoke Lemma~\ref{lem:Bern}, we first derive a uniform bound on
$\|\bm{S}_k\|_2$. By the sub-multiplicativity of the spectral norm,  we
\begin{align}
\|\bm{S}_k\|_2
&\le
\Big\|\bm{I}_L-\frac{1}{N_r}\bm{G}_k^{\mathsf H}\bm{G}_k\Big\|\,
\|\bm{x}_k\|_2 \nonumber\\ \nonumber
&\le
\Big(1+\frac{1}{N_r}\|\bm{G}_k^{\mathsf H}\bm{G}_k\|\Big)\|\bm{x}_k\|_2,\\
& =  \Big(1+\frac{1}{N_r}\|\bm{G}_k\|^{2}\Big)\|\bm{x}_k\|_2 .
\end{align}
For tall random matrices with i.i.d.\ circularly symmetric complex Gaussian
entries, the Bai--Yin law implies that
$\|\bm{G}_k\|\le 2\sqrt{N_r}$ almost surely for sufficiently large $N_r$.
Therefore,
\begin{equation}
\label{eq:sksurlylem1}
\|\bm{S}_k\|_2 \le (1+  \frac{(2\sqrt{N_r})^2}{N_r})\|\bm{x}_k\|_2 = 5\|\bm{x}_k\|_2,~~ k\in[K].
\end{equation}
{\change
We next evaluate the variance term required by Lemma~\ref{lem:Bern}. From the
definition of $\bm{S}_k$, we obtain
\begin{align}
\mathbb{E}[\bm{S}_k^{\mathsf H}\bm{S}_k]
=
\bm{x}_k^{\mathsf H}
\mathbb{E}\!\left[
\bm{I}_L
-\frac{2}{N_r}\bm{G}_k^{\mathsf H}\bm{G}_k
+\frac{1}{N_r^{2}}\bm{G}_k^{\mathsf H}\bm{G}_k\bm{G}_k^{\mathsf H}\bm{G}_k
\right]
\bm{x}_k .\nonumber
\end{align}
Since $\bm{G}_k^{\mathsf H}\bm{G}_k$ is a complex Wishart matrix, its first two
moments satisfy
$\mathbb{E}[\bm{G}_k^{\mathsf H}\bm{G}_k]=N_r\bm{I}_L$ and
$\mathbb{E}[\bm{G}_k^{\mathsf H}\bm{G}_k\bm{G}_k^{\mathsf H}\bm{G}_k]
=N_r(N_r+L)\bm{I}_L$.
Substituting these identities yields
\begin{equation}
\mathbb{E}[\bm{S}_k^{\mathsf H}\bm{S}_k]
=
\bm{x}_k^{\mathsf H}\Big(\frac{L}{N_r}\bm{I}_L\Big)\bm{x}_k
=
\frac{L}{N_r}\|\bm{x}_k\|_2^{2}.
\end{equation}
Consequently,}
\begin{equation}
\label{eq:vralem1}
\Big\|
\sum\nolimits_{k=1}^{K}\mathbb{E}[\bm{S}_k^{\mathsf H}\bm{S}_k]
\Big\|
\le
\frac{L}{N_r}\sum\nolimits_{k=1}^{K}\|\bm{x}_k\|_2^{2}.
\end{equation}

Applying Lemma~\ref{lem:Bern} with the bounds in \eqref{eq:sksurlylem1} and \eqref{eq:vralem1}, we obtain $\Big\|\sum_{k=1}^{K}\bm{S}_k\Big\|_2 \le \epsilon_1$, with probability at least
\begin{equation}
1-(L+1)\exp\!\left(
-\frac{N_r\epsilon_1^{2}/2}{
L\sum_{k=1}^{K}\|\bm{x}_k\|_2^{2}
+\frac{5\epsilon_1}{3}\max_{k}\|\bm{x}_k\|_2}
\right).
\end{equation}
Let $\delta_1>0$ be given and define
$\gamma:=\sum_{k=1}^{K}\|\bm{x}_k\|_2^{2}$.
A sufficient condition for the above probability to be no smaller than
$1-\delta_1$ is
\begin{equation}
N_r
\ge
\frac{2\gamma L}{\epsilon_1^{2}}
\ln\!\Big(\frac{L+1}{\delta_1}\Big),
\end{equation}
which completes the proof.


\subsection{Proof of Lemma~\ref{lem:Delta2}}\label{sec:lemDelta2}
{\change
Let us define the variable $\bm{Z}_{k} \in  \mathbb{C}^{L\times L}$ for $k\in [K]$ as follows:
\begin{align}
   \bm{Z}_{k}  : = \frac{1}{N_r}\sum\nolimits_{k'=1, k'\neq k}^{K} \bm{G}_{k'}^{\mathsf{H}}\bm{G}_k.
\end{align}
Next, considering the fact that the elements of $\bm{G}_k$ and $\bm{G}_{k'}$ are statistically independent, we have   $  \mathbb{E}[ \bm{Z}_{k}]= \bm{0}, \quad {\rm for}~ k \in [K].$
In order to apply Lemma~\ref{lem:Bern}, we need to bound  the norm of each variable, $\|\bm{Z}_{k}\|_2$, and the variance term,  $\|\mathbb{E}[\bm{Z}_{k}^{\mathsf{H}}\bm{Z}_{k}] \|_2$, which we tackle separately in the following. For the norm, we have
\begin{align}
    \nonumber
    &\|\bm{Z}_{k}\|  \leq \frac{1}{N_r}\|\sum_{k'}\bm{G}_{k'}^{\mathsf{H}}\| \|\bm{G}_k\| \leq  \frac{K-1}{N_r}\|\bm{G}_k\| \| \bm{G}_{k'}\| , \\
    &\leq \frac{K}{N_r}\|\bm{G}_k\|^2  \leq \frac{4N_rK}{N_r} \leq {4}K ,
\end{align}
for $k \in [K]$. Since $\|\bm{G}_k\|$ follows Bai-Yin laws (for tall matrices $N_r \gg L$), which can be bounded by a factor $2\sqrt{N_r}$ almost surely~\cite{vershynin2010introduction}. Now, the variance term can be bounded as
\begin{align}
    \mathbb{E}[\bm{G}_{k}^{\mathsf{H}}\bm{G}_{k'}\bm{G}_{k'}^{\mathsf{H}}\bm{G}_k] = N_r L\bm{I}_L.  
\end{align}
Consequently, it yields
\begin{align}
\nonumber
    \Big\| \mathbb{E}[\bm{Z}_k^{\mathsf{H}}\bm{Z}_k]\Big\|  
    & \leq \frac{(K-1)N_rL}{N_r^2}  <\frac{LK}{N_r} .
\end{align}
Thus, the direct results of applying Lemma~\ref{lem:Bern} to  each variable $\bm{Z}_{k}$  for $\epsilon_2/\sum_{k}\|\bm{x}_k\|\sqrt{K}$  together with union bound give us the following upper bound: 
\begin{align}
    \Big\| \sum\nolimits_{k=1}^K\bm{Z}_{k}\bm{x}_k \Big\|_2 \leq   \Big(\sum_{k}\|\bm{Z}_{k} \|^2\Big)^{1/2}\Big(\sum_k\|\bm{x}_k\|_2^2\Big)^{1/2} \leq \epsilon_2,\nonumber
\end{align}
where the first inequality is the Cauchy–Schwarz inequality, 
with probability no less than
\begin{align}
\nonumber
    1-K(L+1)\exp{\Big(\frac{-N_r\epsilon_2^2\gamma^{-1}}{2LK^2  + \frac{8\epsilon_2}{\sqrt{\gamma}3} }{  
     } \Big)}\geq 1-\delta_2.
\end{align}
By rearranging the expression in terms of $\delta_2$, we obtain 
\begin{align}
\label{eq:NR_final_lower}
     N_r \geq  \frac{2\gamma LK^2}{\epsilon_2^2}\ln{\Big(\frac{K(L+1)}{\delta_2}\Big)},
\end{align}
where  $\delta_2>0$. This concludes the proof of Lemma~\ref{lem:Delta2}. }
\subsection{Proof of Lemma~\ref{lem:Delta3}}\label{sec:lemDelta3}

Let us define the variable $\bm{Q}_{k} \in  \mathbb{C}^{L\times N_r}$ for $k\in [K]$ as:
\begin{align}
   \bm{Q}_{k}:= \frac{1}{N_r}\bm{G}_{k}^{\mathsf{H}}, \quad k\in [K].
\end{align}
Thus, we can check that $\mathbb{E}[\bm{Q}_{k}]=\bm{0}$.  Following the same strategy for the proof in Appendix~\ref{sec:lemDelta2}, we first bound $\|\bm{Q}_k\|_2$ as follows:
\begin{align}
  \nonumber
    \|\bm{Q}_k\| \leq \frac{1}{N_r} \|\bm{G}_k\|  \leq  \frac{2\sqrt{N_t}}{N_r}  = \frac{2}{\sqrt{N_r}},
\end{align}
 for $k\in [K]$. Then, we tackle the variance below:
\begin{align}
     \mathbb{E}[\bm{Q}_{k}^{\mathsf{H}}\bm{Q}_{k}] = \mathbb{E}\Big[\frac{1}{N_r^2}\bm{G}_{k}^{\mathsf{H}}\bm{G}_{k}\Big]  = \frac{L\bm{I}_{N_r}}{N_r^2}.
\end{align}
Hence, we obtain 
\begin{align}
     \bigg\|\sum_{k} \mathbb{E}[\bm{Q}_{k}^{\mathsf{H}}\bm{Q}_{k}] \bigg\| \leq \frac{KL}{N_r^2}\leq \frac{K}{N_r}.
\end{align}
{\change
Similarly, Lemma~\ref{lem:Bern} gives us the following error bound.
\begin{align}
   \Big\|\sum\nolimits_{k=1}^K\bm{Q}_k \bm{z}\Big\|_2 \leq  \Big\|\sum\nolimits_{k=1}^K \bm{Q}_k\Big\|\|\bm{z}\|_2 \leq \epsilon_3,
\end{align}
for $N_r \geq  \frac{8L K \sigma_z^2 }{\epsilon_3^2}\ln{\Big(\frac{L+1}{\delta_3}\Big)}$,  with probability at least $1-\delta_3$. Here, we used the fact that  $\|\bm{z}\|\leq 2\sqrt{L}\sigma_z$ almost surely. As a result, we can conclude the proof of Lemma~\ref{lem:Delta3}.}
 
{\change
\subsection{Proof of Theorem~\ref{cor:new-imperfct}}\label{sec:new-imperfct}

Under imperfect CSI, the distortion term $\Delta\bm{y}$ in
\eqref{eq:imperfect_csi_distortion} can be decomposed into an
interference-type component and a noise-type component, i.e., $\Delta\bm{y} = e_{\rm inter}' + e_{\rm noise}'$. To characterize these two components, define
\begin{subequations}
  \begin{align}
    \bar{\bm{Z}}_k &:= \frac{1}{N_r}\sum\nolimits_{k'=1,\,k'\neq k}^{K}
    \bar{\bm{G}}_{k'}^{\mathsf{H}}\bm{G}_k \in \mathbb{C}^{L\times L},\\
    \bar{\bm{Q}}_k &:= \frac{1}{N_r}\bar{\bm{G}}_{k}^{\mathsf{H}}
    \in \mathbb{C}^{L\times N_r}, \qquad  k\in[K],
\end{align}  
\end{subequations}
results in $e_{\rm inter}' = \sum_k\bar{\bm{Z}}_k$ and $e_{\rm noise}' = \sum_k\bar{\bm{Q}}_k$, 
where $\bar{\bm{G}}_{k} := \frac{1}{N_t}\Delta\bm{H}_k\bm{V}_k
\in \mathbb{C}^{N_r\times L}$ whose entries  are i.i.d. and distributed as
$\mathcal{CN}(0,\xi_k^2)$ with $\xi_k=\sigma_{I,k}/\sigma_{h,k}$.

Since the estimation errors $\Delta\bm{H}_k$ are mutually independent
across nodes and are independent of the true channel matrices
$\bm{H}_k$, it follows that $\mathbb{E}[\bar{\bm{Z}}_k]=\bm{0}$ and $
\mathbb{E}[\bar{\bm{Q}}_k]=\bm{0}.$Hence, the matrix Bernstein inequality in Lemma~\ref{lem:Bern} can be
applied in the same manner as in Lemmas~\ref{lem:Delta2} and
\ref{lem:Delta3} for the random matrices $\bar{\bm{Z}}_k$ and $\bar{\bm{Q}}_k$,
respectively. Following the arguments in Lemma~\ref{lem:Delta2}, we obtain $\|\bar{\bm{Z}}_k\|
\leq
\sum_{k',\,k'\neq k} 4K\,\xi_{k'}
\leq
4K\sum_{k=1}^{K}\xi_k ,$
where the last inequality holds since $\xi_k\geq 0$.
Moreover, the corresponding variance term satisfies $\big\|\mathbb{E}[\bar{\bm{Z}}_k^{\mathsf{H}}\bar{\bm{Z}}_k]\big\|
\leq
{L}\sum_{k=1}^{K}\xi_k^2/N_r$.
As a consequence, for any $\epsilon_4>0$ and $\delta_4\in(0,1)$, $\Big\|\sum_{k=1}^{K}\bar{\bm{Z}}_k\bm{x}_k\Big\|_2 \leq \epsilon_4$
holds with probability at least $1-\delta_4$ provided that
\[
N_r \geq
\frac{2\gamma LK\sum_{k}\xi_k^2}{\epsilon_4^2}
\ln\!\Big(\frac{K(L+1)}{\delta_4}\Big).
\]

Next, by applying the same steps as in Lemma~\ref{lem:Delta3} to the
matrices $\bar{\bm{Q}}_k$, we obtain $\|\bar{\bm{Q}}_k\|\leq \frac{2\xi_k}{\sqrt{N_r}},$ and $\Big\|\sum_{k=1}^{K}\mathbb{E}[\bar{\bm{Q}}_k^{\mathsf{H}}\bar{\bm{Q}}_k]\Big\|
\leq \frac{1}{N_r}\sum_{k=1}^{K}\xi_k^2 .$
Therefore, for any $\epsilon_5>0$ and $\delta_5\in(0,1)$, $\Big\|\sum_{k=1}^{K}\bar{\bm{Q}}_k\bm{z}\Big\|_2 \leq \epsilon_5$
with probability at least $1-\delta_5$, provided that
\[
N_r \geq
\frac{8L\sigma_z^2\sum_{k}\xi_k^2}{\epsilon_5^2}
\ln\!\Big(\frac{L+1}{\delta_5}\Big).
\]

To guarantee that both components of $\Delta\bm{y}$ are bounded and that $\|\Delta\bm{y}\|_2 \leq \epsilon''/2$
holds with probability at least $1-\delta/2$, we combine the above two
bounds using the same union bound arguments as in
\eqref{eq:error_union}--\eqref{eq:proofnr}. This yields the sufficient
condition
\begin{align}
\label{eq:nrinterjadid}
N_r \geq
\frac{8c_{\sigma,\gamma}LK^2\gamma\sigma_z^2\bar{\xi}}{\epsilon''^2}
\ln\!\Big(\frac{4K(L+1)}{\delta}\Big),
\end{align}
where $\bar{\xi}=\sum_{k=1}^{K}\xi_k^2/K$. Finally, by invoking Theorem~\ref{th:NError} and the resultant lower bound in \eqref{eq:nrinterjadid}, while setting $\epsilon''/2$ in
\eqref{eq:Nrnumber}, we obtain
\[
\|\bm{r}-\hat{\bm{r}}\|_2
\leq
\|\bm{y}\|_2+\|\Delta\bm{y}\|_2
\leq
\epsilon''/2+\epsilon''/2
=\epsilon'',
\]
with probability at least $1-\delta$, provided that
\[
N_r \geq
\frac{8c_{\sigma,\gamma}LK^2\gamma\sigma_z^2
\max\{\bar{\xi}/K,1\}}{\epsilon''^2}
\ln\!\Big(\frac{4K(L+1)}{\delta}\Big).
\]
This completes the proof.

}


\bibliographystyle{IEEEtran}
\bibliography{IEEEabrv,Ref2}

\end{document}